\documentclass[letterpaper]{article} 
\usepackage{aaai23}  
\usepackage{times}  
\usepackage{helvet}  
\usepackage{courier}  
\usepackage[hyphens]{url}  
\usepackage{graphicx} 
\usepackage{natbib}  
\usepackage{caption} 
\usepackage{booktabs}       
\usepackage{amsfonts}       
\usepackage{nicefrac}       
\usepackage{microtype}      
\usepackage{xcolor}         
\usepackage{newfloat}
\usepackage{listings}
\usepackage{multirow}
\usepackage{multicol}
\usepackage{subfigure}
\usepackage{mathtools}
\usepackage{amsmath}
\usepackage{bm}
\usepackage{enumitem}
\usepackage{amsthm}
\usepackage{amssymb}
\usepackage{wrapfig}
\usepackage{booktabs}
\usepackage{tcolorbox}
\usepackage{wrapfig}
\usepackage[ruled]{algorithm2e}
\usepackage[framemethod=tikz]{mdframed}
\SetKwInput{KwData}{Input}
\SetKwInput{KwResult}{Output}
\title{MA-GCL: Model Augmentation Tricks for Graph Contrastive Learning}
\author{
    Xumeng Gong\textsuperscript{\rm 1}, 
    Cheng Yang\textsuperscript{\rm 1}\thanks{Corresponding author.}
, 
    Chuan Shi\textsuperscript{\rm 1}
}

\affiliations{
    \textsuperscript{\rm 1} Beijing University of Posts and Telecommunications\\
    Xumeng1141@bupt.edu.cn,yangcheng@bupt.edu.cn,  shichuan@bupt.edu.cn
}

\usepackage{bibentry}

\begin{document}

\maketitle

\begin{abstract}
Contrastive learning (CL), which can extract the information shared between different contrastive views, has become a popular paradigm for vision representation learning. Inspired by the success in computer vision, recent work introduces CL into graph modeling, dubbed as graph contrastive learning (GCL). However, generating contrastive views in graphs is more challenging than that in images, since we have little prior knowledge on how to significantly augment a graph without changing its labels. We argue that typical data augmentation techniques (\textit{e.g.}, edge dropping) in GCL cannot generate diverse enough contrastive views to filter out noises. Moreover, previous GCL methods employ two view encoders with exactly the same neural architecture and tied parameters, which further harms the diversity of augmented views. To address this limitation, we propose a novel paradigm named model augmented GCL (MA-GCL), which will focus on manipulating the architectures of view encoders instead of perturbing graph inputs. Specifically, we present three easy-to-implement model augmentation tricks for GCL, namely \textit{asymmetric}, \textit{random} and \textit{shuffling}, which can respectively help alleviate high-frequency noises, enrich training instances and bring safer augmentations. All three tricks are compatible with typical data augmentations. Experimental results show that MA-GCL can achieve state-of-the-art performance on node classification benchmarks by applying the three tricks on a simple base model. Extensive studies also validate our motivation and the effectiveness of each trick. (Code, data and appendix are available at \url{https://github.com/GXM1141/MA-GCL}. )
\end{abstract}
\section{Introduction}

Contrastive learning (CL) has emerged as a promising paradigm for unsupervised vision representation learning~\cite{chen2020simple,he2020momentum,grill2020bootstrap, van2018representation, hjelm2018learning}. In short, typical CL methods will generate two views of the same sample by data augmentations, and then maximize the similarity between their encoded representations. In this way, CL can extract the information shared between different views~\cite{tian2020makes}, and thereby alleviate the task-irrelevant noises that only appear in a single view. Recently, CL is also introduced to graph domain for unsupervised graph representation learning, dubbed as graph contrastive learning (GCL)~\cite{you2020graph, zhu2021graph, hassani2020contrastive, zhu2020deep,qiu2020gcc, suresh2021adversarial}. GCL approaches have achieved competitive performance on various graph benchmarks compared to the counterpart trained with ground-truth labels.

However, generating data augmentations on graphs is more challenging than that on images.
Theories on contrastive learning~\cite{tian2020makes} show that good contrastive views should be diverse while keeping the task-relevant information untouched. In computer vision area, we have strong prior knowledge on how to significantly augment an image without changing its labels (\textit{e.g.}, rotation or shift). While in graph learning area, dropping a single edge could possibly destroy a key connection (\textit{e.g.,} an important chemical bond) related to downstream tasks. Nevertheless, most existing graph data augmentation (GDA) techniques (\textit{e.g.,} node or edge dropping) choose to randomly perturb the graph topology~\cite{velickovic2019deep,you2020graph, hassani2020contrastive, zhu2020deep,qiu2020gcc} as contrastive views. 

We argue that existing GDA techniques are in a dilemma: the perturbations that preserve sufficient task-relevant information cannot generate diverse enough augmentations to filter out noises. Moreover, previous GCL methods employ two view encoders with exactly the same neural architecture and tied parameters\footnote{Though some work~\cite{thakoor2021bootstrapped,xia2022simgrace} employs momentum update and hence makes two view encoders not entirely the same, we argue that such difference is not enough.}, which further harms the diversity of augmented views. Though some recent GCL methods generate augmentations in a heuristic~\cite{zhu2021contrastive} or adversarial~\cite{you2021graph,suresh2021adversarial} way, they still use two view encoders with identical architectures and fail to fully address this problem. Besides, a very recent work~\cite{xia2022simgrace} proposes to perturb model parameters instead of graph inputs as augmentations. Note that the function of encoded semantics with respect to model parameters is rather complex. Hence the parameter perturbation drawn from a simple distribution (\textit{e.g.}, Gaussian distribution) may harm the semantics in encoded representations. We will empirically validate our motivation in experiments. 

In this work, we propose a novel paradigm for GCL to address the above limitations, named as MA-GCL (\textbf{M}odel \textbf{A}ugmented \textbf{G}raph \textbf{C}ontrastive \textbf{L}earning). We interpret graph neural network (GNN) encoders as a composition of propagation operators (\textit{i.e.}, graph filters) and transformation operators (\textit{i.e.}, weight matrix and non-linearity). Note that previous GCL methods adopt fixed and identical GNNs as view encoders, \textit{i.e.}, the numbers of propagation and transformation operators, as well as their permutations, are fixed and identical for encoding different contrastive views. While in this work, we focus on perturbing the neural architectures of view encoders as contrastive views, and present three model augmentation tricks: (1) \textbf{Asymmetric strategy}: we will use two contrastive encoders with different numbers of propagation operators. We theoretically prove that this strategy can help alleviate high-frequency noises. (2) \textbf{Random strategy}: we will randomly vary the number of propagation operators in every epoch. The intuition behind is that varying the propagation depth can enrich the diversity of training instances, and hence help predict downstream task. (3) \textbf{Shuffling strategy}: we will shuffle the permutations of propagation and transformation operators in two view encoders. The intuition behind is that shuffling the order of operators will not change the semantics of an input graph, but will perturb the encoded representations as safer augmentations. All three tricks are simple and compatible with typical data augmentations.

We conduct experiments on six graph benchmarks to show the superiority of MA-GCL. For the implementation of MA-GCL, we combine the three tricks and apply them on a simple base model, which can be seen as a simplified version of GRACE~\cite{zhu2020deep} and performs worse than GRACE and recent state-of-the-art (SOTA) methods. Experimental results show that MA-GCL achieves SOTA performance on 5 out of 6 benchmarks, and the relative improvement against best performed baseline can go up to $2.7\%$. Extensive experiments further show that all three tricks will contribute to the overall improvement, and the \textit{Asymmetric strategy} is the most effective one among the three.

Our contributions are as follows:

(1) We highlight a key limitation in most GCL methods that the graph augmentations are not diverse enough to filter out noises. To overcome this limitation, we propose a novel paradigm named model augmentation for GCL, which will focus on perturbing the architectures of GNN encoders instead of graph inputs or model parameters.

(2) We present three effective model augmentation tricks for GCL, namely \textit{asymmetric}, \textit{random} and \textit{shuffling}, which can respectively help alleviate high-frequency noises, enrich training instances and bring safer augmentations. All three tricks are simple, easy-to-implement and compatible with typical data augmentations.

(3) Experimental results show that, compared with recent SOTA GCL methods, MA-GCL can achieve SOTA performance on 5 out of 6 graph benchmarks by applying the three tricks on a simple base model. Extensive studies also validate our motivation and the effectiveness of each strategy.






\section{Related Works}

\paragraph{Graph Contrastive Learning} Recently, many GCL methods were proposed ~\cite{sun2020infograph, wang2019deep, zhu2020deep,qiu2020gcc} for unsupervised node/graph representation learning, and achieved the SOTA performance on this task.  We roughly categorize previous GCL methods into two groups, depending on how they generate contrastive views: (1) The first group usually generates views based on different scales of substructures. InfoGraph~\cite{sun2020infograph} treats the graph structure of different scales (\textit{e.g.,}, nodes, edges, triangles) as contrastive views to learn graph-level representations. GCC~\cite{qiu2020gcc} introduces InfoNCE to large-scale graph pre-training by adopting different sub-graphs as contrastive views. (2) The second group usually generates views by applying data augmentation on graph inputs. GRACE~\cite{zhu2020deep} and GraphCL~\cite{you2020graph} randomly perturb the graph structures (\textit{e.g.,}, node \& edge dropping, graph sampling) to build contrastive views. MVGRL ~\cite{hassani2020contrastive} and MV-CGC~\cite{yuan2021semi} further employ graph diffusion to generate augmentation views of original graphs. BGRL ~\cite{thakoor2021bootstrapped} performs GCL without negative pairs inspired by BYOL~\cite{grill2020bootstrap}. ARIEL~\cite{feng2022adversarial} further introduces an extra adversarial view to GCL. Recent methods propose to generate views in adaptive ways. GCA~\cite{zhu2021graph} designs augmentation schemes based on different graph property (\textit{e.g.,}, node degree, eigenvector, PageRank). DiGCL~\cite{tong2021directed} uses Laplacian perturbation without changing directed graph structure. JOAO~\cite{you2021graph} and AD-GCL~\cite{suresh2021adversarial} automatically select graph augmentations in adversarial ways. InfoGCL~\cite{xu2021infogcl} uses task labels to select optimal views by information bottleneck principle. \cite{ssllpa_icml22} proposes label-preserving augmentations (LPA) to improve GCL. RGCL~\cite{rgcl_icml22} creates rationale-aware views for CL. Besides the two groups above, there are some approaches generating views based on mathematical interpretations or parameter perturbations. COLES~\cite{zhu2021contrastive} realizes better negative sampling strategy and extends Laplacian Eigenmaps for GCL. DSGC~\cite{yang2022dual} conducts CL among views from hyperbolic and Euclidean spaces. GASSL~\cite{yang2021graph} generates views by adding perturbations to both input features and hidden layers. SimGRACE~\cite{xia2022simgrace} adds Gaussian noises to model parameters as contrastive views without data augmentation.  \cite{yu2022graph} adds uniform noises to the embedding space.
Note that all these GCL methods employ two GNNs with the same fixed architecture as view encoders. In this work, we propose a novel paradigm for GCL, where the architectures of view encoders in MA-GCL are different and randomized. 

\paragraph{Relevant Theories and Techniques} In terms of the motivation of MA-GCL, the principle of keeping more task-relevant information and reducing more noise, has been studied through Information Bottleneck~\cite{tishby1999information} theory and minimal sufficient statistics ~\cite{soatto2014visual} in computer vision~\cite{tian2020makes}. Some recent GCL methods~\cite{suresh2021adversarial,xu2021infogcl} are also developed with this principle. In terms of the randomness of MA-GCL, the idea of using randomized GNN architectures is quite different from Graph Random Neural Networks (GRAND)~\cite{feng2020graph,feng2022grand+}. GRAND designs multiple channels of GNNs and ensembles them for prediction, where the input features are randomized in each channel. Thus the neural architecture of GRAND is fixed, and the aim of GRAND is to address the over-smoothing and non-robustness issues in semi-supervised learning.
\section{Notations and Preliminaries}
\paragraph{Notations} Let $\mathcal{G} = (\mathcal{V}, \mathcal{E})$ be a graph, where $\mathcal{V}=\{1,...,|\mathcal{V}|\}$ is the set of $|\mathcal{V}|$ vertices and $\mathcal{E}\subseteq{\mathcal{V}\times\mathcal{V}}$ is the set of $|\mathcal{E}|$ edges between vertices. $\bm{A} \in \{0, 1\}^{|\mathcal{V}|\times |\mathcal{V}|}$ denotes the adjacency matrix of $\mathcal{G}$ with self-loops, where $\bm{A}_{i,j}=1$ if $(v_i,v_j)\in \mathcal{E}$ or $i=j$. Each node $v_i$ is associated with a feature vector $\bm{x}_i \in \bm{X} \in  \mathbb{R}^{|\mathcal{V}|\times d}$, where $d$ is the feature dimension. 

\paragraph{Graph Neural Networks (GNNs)}
GNNs ~\cite{kipf2016semi,velickovic2017graph,zhou2020graph} generalize deep learning techniques into graphs for node encoding. Specifically, GNNs will stack multiple propagation layers and transformation layers, and then apply them to the raw features $\bm{X}$. Here we denote the operators of propagation layer and the transformation layer as $g$ and $h$, respectively:

\begin{equation}
    g (\bm{Z}; \bm{F}) = \bm{F}\bm{Z},\; 
    h (\bm{Z}; \bm{W})=\sigma(\bm{Z}\bm{W}),
\end{equation}
where $\bm{Z}\in  \mathbb{R}^{|\mathcal{V}|\times d_Z}$ is node embeddings, $\bm{F}\in  \mathbb{R}^{|\mathcal{V}|\times |\mathcal{V}|}$ is the graph filter matrix, $\bm{W}\in  \mathbb{R}^{d_Z\times d_O}$ denotes the weight matrix of linear transformation, and $\sigma$ is the nonlinear function (\textit{e.g.,} ReLU). Graph filter $\bm{F}$ is a constant matrix based on the adjacency matrix $\bm{A}$, while $\bm{W}$ is trainable parameters. The output of $g$ has the same shape with input matrix $\bm{Z}$, while the output of $h$ is a ${|\mathcal{V}|\times d_O}$ dimensional representation matrix. With the help of the two operators, node representations can be propagated to their neighbors, and mapped to a new embedding space by linear and nonlinear transformations. 

Now we can formalize GNNs as the composition of multiple $g$ and $h$ operators. For example, a $L$-layer Graph Convolutional Network (GCN)~\cite{kipf2016semi} and a $L$-layer SGC~\cite{wu2019simplifying} can be written as
\begin{equation}
\begin{aligned}
    &{GCN}(\bm{X})=h_L\circ g\circ h_{L-1}\circ g\circ {\cdots}\circ h_1\circ g(\bm{X}),\; \\
    &{SGC}(\bm{X})=h\circ g^{[L]}(\bm{X}),
    \end{aligned}
\end{equation}
where $\circ$ denotes the composition of two operators, $h_i$ denotes the $i$-th transformation layer of GCN, and $g^{[L]}$ denotes the composition of $L$ operators of $g$. The graph filters in GCN and SGC are computed by normalizing the adjacency matrix $\bm{A}$ as: $\bm{F}=\bm{D}^{-\frac{1}{2}}\bm{A}\bm{D}^{-\frac{1}{2}}$, where $\bm{D}$ is the degree matrix. Following many previous GNNs~\cite{chen2020simple, feng2020graph, cui2020adaptive}, we define the graph filter as $\bm{F}=(1-\pi)\bm{I}+\pi\bm{D}^{-\frac{1}{2}}\bm{A}\bm{D}^{-\frac{1}{2}}$, where $\pi\in (0,1)$ and $\bm{I}$ is the identity matrix. In our experiments, we fix $\pi=0.5$.


\paragraph {Graph Contrastive Learning (GCL)}
Given a graph dataset $\mathcal{D}$ of observations, the purpose of GCL is to learn the representations of graphs or nodes in an unsupervised way~\cite{you2020graph, zhu2021graph, hassani2020contrastive, zhu2020deep}. A typical GCL model consists of three key modules~\cite{you2020graph, zhu2021graph, suresh2021adversarial}: graph data augmentation (GDA), view encoders, and contrastive loss. In particular, GDA (\textit{e.g.,} node or edge dropping) will generate different augmentations of the same observation; view encoders (\textit{e.g.,} GCN~\cite{kipf2016semi}) will transform the augmented graphs into view representations; contrastive loss (\textit{e.g.,} InfoNCE) will maximize the consistency between different views, and can identify the invariant parts of view representations.


Formally, for each observation $s$ (\textit{e.g.,} a node or a graph), GDA will generate a pair of augmented views $(a(s), a'(s))$. Then GCL uses two view encoder functions $(f,f')$ to map the augmentation pairs to corresponding representations $(\bf{z}, \bf{z}')$=$(f(a(s)), f'(a'(s))$. Typically, the two view encoders in GCL have exactly the same neural architecture~\cite{zhu2021graph, hassani2020contrastive, zhu2020deep}, \textit{i.e.,} $f=f'$. Finally, InfoNCE loss~\cite{hjelm2018learning} will be minimized to enforce a similar representation across positive pairs:
{\small
\begin{equation}
\mathcal{L}=-\sum\limits_{i=1}^N \log {\frac{\exp{(-|\bf{z}_i-\bf{z}_i'|^2/2)}}{\exp{(-|\bf{z}_i-\bf{z}_i'|^2/2)}+\sum_{j\neq i} \exp{(-|\bf{z}_i-\bf{z}_j|^2/2)}}},
\label{eq:clloss}
\end{equation}
}
where $(\bf{z}_i,\bf{z}_i')$ and  $(\bf{z}_i,\bf{z}_j)$ are the positive and negative pairs of view representations, respectively.  As shown in ~\cite{jing2021understanding}, the above square loss form is equivalent to the typical cosine similarity form of normalized embeddings.


\section{Methodology}
To address the limitations of existing GCL methods, we will focus on manipulating the neural architectures of view encoders as model augmentations. In this section, we will first present three strategies and their benefits separately. Then we will illustrate the overall algorithm by incorporating the three tricks to a simple base model.



\subsection{Asymmetric Strategy}
\label{Asymmetry}
\begin{figure}[] 
\vspace{-0.5cm}
\centering
    \subfigure[Symmetric]{
        \label{vn1}
        \includegraphics[scale=0.33]{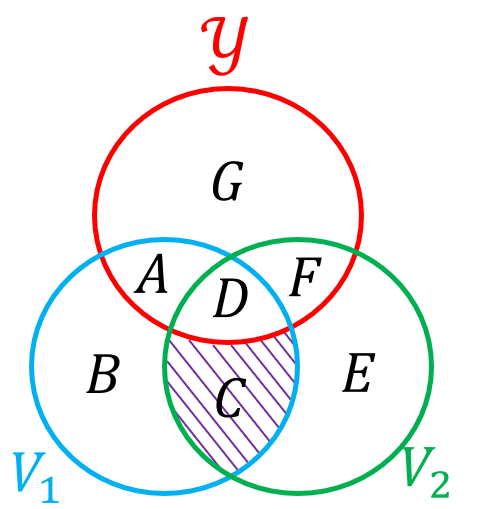}
    }
    \hspace{-1cm}
    \hfill
    \subfigure[Asymmetric]{
        \includegraphics[scale=0.33]{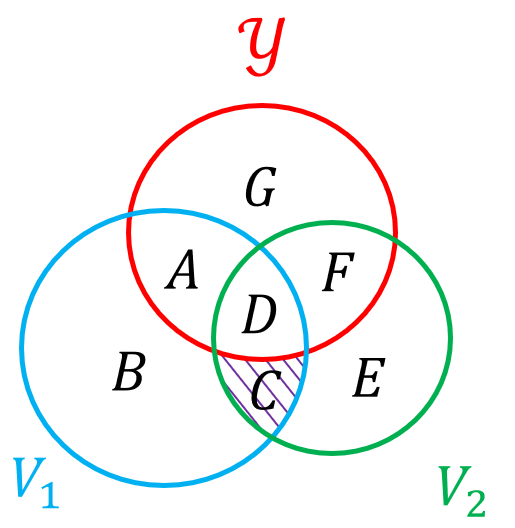}
        \label{vn2}
    }
\caption{An illustration of our motivation about the asymmetric strategy. Here $V_1, V_2$ and $\mathcal{Y}$ represent the information of two views and downstream tasks, respectively. The  strategy can push the two views away from each other, and the task-irrelevant noises in GCL (area C) can be alleviated.}
\label{fig:vn}
\end{figure}

\paragraph{One-sentence Summary} Using encoders with shared parameters but different numbers of propagation layers can alleviate high-frequency noises.
\paragraph{Main Idea} 
Contrastive learning (CL) can extract the information shared between different views~\cite{tian2020makes}, and thereby filter out the task-irrelevant noises that only appear in a single view. As shown in Fig.~\ref{fig:vn} where the scale of each area denotes the amount of information, the learned representations of CL include both task-relevant information (area D) and task-irrelevant noises (area C). Intuitively, the two views should not be too far (limited information of D) or too close (too much noises of C). We argue that the two views in previous GCL methods are too close to each other: (1) Typical GDA techniques (\textit{e.g.,} edge or node dropping) cannot generate diverse enough augmentations while keeping the task-relevant information untouched; (2) The two view encoders have exactly the same neural architecture and tied parameters, which strengthens the closeness between views.

To address this problem, we propose to use asymmetric view encoders with shared parameters but different numbers of propagation layers. As shown in Fig.~\ref{vn2}, we can push the two views away from each other by different numbers of propagation layers, and shared parameters can make sure that the distance between them will not be too far. In this way, the noises in GCL (area C) can be  alleviated. Note that the numbers of $h$ and $g$ operators are totally detached. For implementation, we employ two GNN encoders with the same number of $h$ operators but different numbers of $g$ operators. Now we will show that our implementation can help filter out high-frequency noises.

\paragraph{Benefits} For simplicity, we ignore the non-linear transformations in GNN encoders to demonstrate our benefits. Without non-linear functions, the weight matrices in $h$ operators collapse to a single weight matrix $\bm{W}$. Then we have the embeddings of two encoders as $\bm{Z}=\bm{F}^{L}\bm{X}\bm{W},\; \bm{Z}'=\bm{F}^{L'}\bm{X}\bm{W}$, where $L,L'$ are the numbers of $g$ operators in two view encoders, and $L<L'$. 

Assume that graph filter $\bm{F}$ can be factorized to $\bm{U}\Lambda \bm{U}^T$ by eigenvector decomposition, where $\bm{U}$ is the unitary matrix of eigenvectors and $\Lambda=\mbox{diag}(\lambda_1,\lambda_2,\dots,\lambda_{|\mathcal{V}|})$ is the diagonal matrix of eigenvalues $\lambda_1\geq\lambda_2\geq\dots \lambda_{|\mathcal{V}|}$. A larger eigenvalue of $\bm{F}$ corresponds to a smaller eigenvalue of graph Laplacian matrix, which is often recognized as more important information for downstream tasks~\cite{nt2019revisiting}. Here we focus on analyzing graph signals and assume one-hot node features $\bm{X}=\bm{I}$. Then the embeddings of two views can be written as $\bm{Z}=\bm{U}\Lambda^{L}\bm{U}^T\bm{W},\; \bm{Z}'=\bm{U}\Lambda^{L'}\bm{U}^T\bm{W}$, where $\bm{W}\in \mathbb{R}^{|\mathcal{V}|\times d_O}$.

Contrastive loss aims to minimize the distance between embeddings of a positive pair. Following this idea, we focus on the numerator term in Eq.~(\ref{eq:clloss}) and rewrite the loss as
{\small
\begin{equation}
\begin{aligned}
    \min_{\bm{W}} \sum_{i=1}^{|\mathcal{V}|} |\bm{z}_i-\bm{z}_i'|^2 &=\min_{\bm{W}} \mbox{tr}((\bm{Z}-\bm{Z}')(\bm{Z}-\bm{Z}')^T), \\
    &\quad \mbox{subject to}\quad  \bm{W}^T\bm{W}=\bm{I},
\label{eq:asym}
\end{aligned}
\end{equation}}
where $\mbox{tr}(\cdot)$ denotes matrix trace, and the condition $\bm{W}^T\bm{W}=\bm{I}$ is to avoid trivial all-zero solutions. 

\newtheorem{thm}{\bf Theorem }
\begin{thm}\label{thm1}
The optimal $\bm{W}$ of the contrastive learning loss in Eq.~(\ref{eq:asym}) is
\begin{equation}
\bm{W}^*=\bm{U}[k_1,k_2,\dots k_{d_O}],
\end{equation}
where $1\leq k_1<k_2<\dots k_{d_O}\leq |\mathcal{V}|$ are the best $k$s that minimize $(\lambda_k^L-\lambda_k^{L'})^2$, and $U[k_1,k_2,\dots k_{d_O}]$ denotes the corresponding columns of $\bm{U}$.
\end{thm} 

The detailed proof of Theorem~\ref{thm1} is presented in Appendix.\textbf{A}. Now we will see which $k$s have the smallest $(\lambda_k^L-\lambda_k^{L'})^2$. The function $\min_\lambda(\lambda^L-\lambda^{L'})^2$ prefers $\lambda\rightarrow 0$ or $\lambda\rightarrow 1$. Taking the graph filter $\bm{F}=\frac{1}{2}\bm{I}+\frac{1}{2}\bm{D}^{-\frac{1}{2}}\bm{A}\bm{D}^{-\frac{1}{2}}$ used in our experiments as an example, the eigenvalues of $\bm{F}$ fall in range $[0,1]$. As shown in ~\cite{cui2020adaptive}, for popular graph datasets such as Cora and Citeseer~\cite{kipf2016semi}, the eigenvalues of $\bm{F}$ will fall in range $[\frac{1}{4},1]$ in practice. Besides, we also find that the filter $\bm{F}$ of these popular graph data has many eigenvalues close to $1$, \textit{e.g.,} the $100$-th largest eigenvalue of $\bm{F}$ in Cora is still larger than $0.998$. Therefore, $\bm{U}[k_1,k_2,\dots k_{d_O}]$ is likely to be the first $d_O$ columns of $\bm{U}$ in practice. Then the representations of view encoder are 
\begin{equation}
    \bm{Z}=\bm{U}\Lambda^{L}\bm{U}^T\bm{W}=\bm{U}\mbox{diag}(\lambda_1^L,\lambda_2^L,\dots, \lambda_{d_O}^L,0,0,\dots,0),
\end{equation}
which can perfectly filter out the high-frequency noises. 

In contrast, if we adopt identical view encoders with $L=L'$, then $\bm{Z}-\bm{Z}'=\bm{U}(\Lambda^{L}-\Lambda^{L'})\bm{U}^T\bm{W}=0$. Thus, the perturbations of GDA are no longer minor terms and have to be considered. Assume that the influence of GDA on eigenvalues is $\epsilon\in \mathbb{R}^{|\mathcal{V}|}$, we have $(\Lambda+\mbox{diag}(\epsilon))^{L}-\Lambda^{L}\approx L\mbox{diag}(\epsilon)\Lambda^{L-1}$. Thus the optimal $\bm{W}^*$ is determined by both filter $\bm{F}$ and the perturbations of GDA, which will probably incorporate high-frequency noises into learned representations.


\subsection{Random Strategy}

\paragraph{One-sentence Summary} Randomly varying the number of propagation operators in view encoders at every epoch can help enrich training samples.

\paragraph{Main Idea} Without loss of generality, we illustrate our idea based on SGC~\cite{wu2019simplifying} encoder, \textit{i.e.,} $f(\bm{X})=h\circ g^{[L]}(\bm{X})$.  During the training process of GCL, we propose to vary the number of operator $g$ in the encoder $f$. Specifically, we randomly sample $\hat{L}\sim \mbox{Uniform}(lower, upper)$ in every epoch, instead of employing a fixed $L$. The intuition behind is that varying the propagation depth can enrich the diversity of training instances, and hence help predict downstream tasks. Now we will present the benefits more formally.

\paragraph{Benefits} As a GNN encoder with $L$ propagation operators of $g$, $f$ will compute node representation $\bm{Z}_v \in \bm{Z}$ for every node $v$ by aggregating $L$-hop neighbors of node $v$. In other words, $f$ is equivalent to a function applied to the local computation tree~\cite{nt2019revisiting} rooted at node $v$, with a depth of $L$.

Let $\mathcal{S}=\{v\}_{v=1}^{|\mathcal{V}|}$ be the training set of GCL. 
Then we can replace the observations in $\mathcal{S}$ from the individual node $v$ to its $L$-hop computation tree $t_{v}^L$, and rewrite the training set as $\tilde{\mathcal{S}}^L=\{t_v^L\}_{v=1}^{|\mathcal{V}|}$. 
When the number of $g$ operators is randomly chosen from $\hat{L}\sim \mbox{Uniform}(lower, upper)$ in every epoch, we can hypothesize that the set of samples changes to ${\tilde{\mathcal{S}}=\bigcup_{\hat{L}\in[lower,upper]}{\tilde{\mathcal{S}}^{\hat{L}}}}$. 
Therefore, the training set of GCL gets enlarged by several times. In fact, if we further take GDA (\textit{e.g.,} edge dropping) into consideration, the training set will involve the computation trees and their sub-trees. 
There could be an even wider gap (\textit{e.g.,} exponentially) between the set capacity with random $\hat{L}$ and that with fixed $L$.

\subsection{Shuffling Strategy}
\paragraph{One-sentence Summary} Shuffling the permutation of propagation and transformation operators in every epoch brings safer augmentations.

\paragraph{Main Idea} 
Existing GDA techniques usually randomly perturb the graph topology and will take the risk of destroying key connections (\textit{e.g.,} dropping an important chemical bond) related to downstream tasks. To address this problem, we propose to use different permutations of operators $g$ and $h$ in two view encoders as safer augmentations. Formally, if view encoder $f$ has $L$ operators of $g$ and $N$ operators of $h$, then $f$ can be written as
\begin{equation}\label{eq:encoder}
f(\bm{X})= h_N \circ g^{[K_{N}]}\circ h_{N-1} \circ \dots g^{[K_2]}\circ h_1 \circ g^{[K_1]}(\bm{X}),
\end{equation}
where $K_1\dots K_N\geq 0$ and $\sum_{i=1}^N K_i=L$. Then we will use a different set of $K_1',K_2'\dots K_N'\geq 0$ with $\sum_{i=1}^N K_i'=L$ for another view encoder $f'$. The intuition behind is that shuffling the order of propagation and transformation operators will not change the semantics of an input graph, but will perturb the encoded representations as safer augmentations. Now we will discuss more about the relationship between our strategy and previous GDA techniques.



\paragraph{Benefits} As shown in {\cite{wang2020understanding}}, contrastive learning includes an important step of alignment, whose metric is as follows:

\begin{equation}
\mathcal{L}_{align} \triangleq \mathbb{E}_{x\sim \mathcal{D}, x'\sim p_{pos}(x)}[||f_\theta(x)-f'_{\theta'}(x')||_{2}],
\label{eq:align}
\end{equation}
where $p_{pos}(x)$ is the distribution of data augmentation, and $\theta,\theta'$ denote the learnable parameters in encoders $f,f'$.

Previous GCL methods based on data augmentation~\cite{zhu2021graph,you2020graph} can be interpreted as $f'=f$ and $\theta'=\theta$ in Eq.~(\ref{eq:align}). Those based on perturbing model parameters~\cite{yang2021graph,xia2022simgrace} can be interpreted as $f'=f$, $x'=x$ and $\theta'=\theta+\mathcal{N}$ where $\mathcal{N}$ is random perturbation noise (\textit{e.g.}, drawn from a Gaussian distribution). Note that the function of encoded semantics with respect to model parameters is rather complex. Hence the parameter perturbation drawn from a simple distribution may also harm the semantics in encoded representations. While in this strategy, we have $\theta'=\theta$ and $x'=x$. Note that our $f$ and $f'$ will be the same if we ignore all the non-linear transformations. Therefore, $f'$ can provide safer augmentations by utilizing the perturbations brought by non-linear transformations.

\subsection{Implementation of MA-GCL}
\label{sec:implement}
Now we will present our MA-GCL by applying the three strategies on a simple base model. The base model adopts random edge/feature dropping, graph filter $\bm{F}=\frac{1}{2}\bm{I}+\frac{1}{2}\bm{D}^{-\frac{1}{2}}\bm{A}\bm{D}^{-\frac{1}{2}}$, $2$-layer embedding projector, and InfoNCE loss for learning. Note that the base model can be seen as a simplified version of GRACE~\cite{zhu2020deep} with only intra-view modeling. The pseudo code is shown in Alg. {\ref{Alg_magcl}}. After the training phase, we will use a fixed encoder architecture and drop the embedding projector for evaluation: similar to the architecture of GCN, we set $K_1=K_2=\dots K_{N}=K$, where $K\in\{1,2\}$ is a hyper-parameter. 

    \begin{algorithm}
    
    \caption{Implementation of MA-GCL}
    \KwData{Adjacency matrix $\bm{A}$; feature matrix $\bm{X}$; GDA distribution $\mathcal{T}$; Random range $[low, high]$;\\
    \ \ \ \ \ \ \ \ \ \ \ $N$ transformation operators $h_1,h_2\dots h_N$;\\
    \ \ \ \ \ \ \ \ \ \ \ Embedding projector proj($\cdot$) \;\\}
    \KwResult{$N$ learned operators $h_1,h_2\dots h_N$.}
    
    \For{every $epoch$}{
    	Draw two graph augmentations: $a,a' \sim \mathcal{T}$, $(\bm{A}_1, \bm{X}_1) = a(\bm{A}, \bm{X}),(\bm{A}_2, \bm{X}_2) = a'(\bm{A}, \bm{X})$\;
        Draw $K_i,K_i'$ from Uniform$(low, high)$ for $i=1\dots N$\;
        Assure $L=\sum_{i=1}^N K_i\neq \sum_{i=1}^N K_i'=L'$, and $\forall$ $i$ $K_i\neq K_i'$\;
        Calculate graph filters in $g_1,g_2$:\ $\bm{F}_1=\frac{1}{2}\bm{I}+\frac{1}{2}\bm{D_1}^{-\frac{1}{2}}\bm{A_1}\bm{D_1}^{-\frac{1}{2}}$,  $\bm{F}_2=\frac{1}{2}\bm{I}+\frac{1}{2}\bm{D_2}^{-\frac{1}{2}}\bm{A_2}\bm{D_2}^{-\frac{1}{2}}$;\\
        Augmented view encoders: $f_1=h_N \circ g_1^{[K_{N}]} \dots  h_1 \circ g_1^{[K_1]}$, $f_2=h_N \circ g_2^{[K'_{N}]} \dots  h_1 \circ g_2^{[K'_1]}$;\\
        View encoding: $\bm{Z}_{1}=f_{1}(\bm{X_{1}}),\bm{Z}_{2}=f_{2}(\bm{X_{2}})$\\
        Update parameters by contrastive loss $\mathcal{L}(\mbox{proj}(\bm{Z}_1), \mbox{proj}(\bm{Z}_2))$;\\
    }
    \label{Alg_magcl}
    \end{algorithm}

\section{Experiments}
\begin{table*}[htbp]

  \centering
  \footnotesize
    \begin{tabular}{cc|ccc|cccccccccccc|ccc}
    \toprule[2pt]
    \multicolumn{3}{c|}{Datasets} & \multicolumn{2}{c}{Cora} & \multicolumn{2}{c}{CiteSeer} & \multicolumn{2}{c}{PubMed} & \multicolumn{2}{c}{Coauthor-CS} & \multicolumn{2}{c}{Amazon-C} & \multicolumn{2}{c}{Amazon-P} & \multicolumn{2}{|c}{Avg. Acc.}& \multicolumn{1}{c}{Avg. Rank}\\
    \midrule
     \multicolumn{3}{c|}{GCN} & \multicolumn{2}{c}{82.5 ± 0.4} & \multicolumn{2}{c}{71.2 ± 0.3} & \multicolumn{2}{c}{79.2 ± 0.3} & \multicolumn{2}{c}{ 93.03 ± 0.3} & \multicolumn{2}{c}{86.51 ± 0.5} & \multicolumn{2}{c}{ 92.42 ± 0.2} & \multicolumn{2}{|c}{ }& \multicolumn{1}{c}{}\\
    \multicolumn{3}{c|}{GAT} & \multicolumn{2}{c}{83.0 ± 0.7} & \multicolumn{2}{c}{72.5 ± 0.7} & \multicolumn{2}{c}{79.0 ± 0.3} & \multicolumn{2}{c}{92.31 ± 0.2} & \multicolumn{2}{c}{86.93 ± 0.3} & \multicolumn{2}{c}{92.56 ± 0.4} & \multicolumn{2}{|c}{ -}& \multicolumn{1}{c}{ -}\\
    \multicolumn{3}{c|}{InfoGCL} & \multicolumn{2}{c}{83.5 ± 0.3} & \multicolumn{2}{c}{73.5 ± 0.4} & \multicolumn{2}{c}{79.1 ± 0.2} & \multicolumn{2}{c}{-} & \multicolumn{2}{c}{-} & \multicolumn{2}{c}{-} & \multicolumn{2}{|c}{ }& \multicolumn{1}{c}{ }\\
    \midrule
    \multicolumn{3}{c|}{DGI} & \multicolumn{2}{c}{82.3 ± 0.6} & \multicolumn{2}{c}{71.8 ± 0.7} & \multicolumn{2}{c}{76.8 ± 0.3} & \multicolumn{2}{c}{ 92.15 ± 0.6} & \multicolumn{2}{c}{83.95 ± 0.5} & \multicolumn{2}{c}{91.61 ± 0.2}  & \multicolumn{2}{|c}{83.10} & \multicolumn{1}{c}{8.5}\\
    \multicolumn{3}{c|}{ GRACE } & \multicolumn{2}{c}{81.7 ± 0.4} & \multicolumn{2}{c}{71.5 ± 0.5} & \multicolumn{2}{c}{ 80.7 ± 0.4} & \multicolumn{2}{c}{92.93 ± 0.0} & \multicolumn{2}{c}{87.46 ± 0.2} & \multicolumn{2}{c}{92.15 ± 0.2} & \multicolumn{2}{|c}{84.44} & \multicolumn{1}{c}{6.5} \\
    \multicolumn{3}{c|}{MVGRL} & \multicolumn{2}{c}{83.4 ± 0.3} & \multicolumn{2}{c}{73.0 ± 0.3} & \multicolumn{2}{c}{ 80.1 ± 0.6} & \multicolumn{2}{c}{ 92.11 ± 0.1} & \multicolumn{2}{c}{87.52 ± 0.1} & \multicolumn{2}{c}{ 91.74 ± 0.0} & \multicolumn{2}{|c}{84.63}  & \multicolumn{1}{c}{6.5}\\
    \multicolumn{3}{c|}{BGRL} & \multicolumn{2}{c}{81.7 ± 0.5} & \multicolumn{2}{c}{72.1 ± 0.5} & \multicolumn{2}{c}{80.2 ± 0.4} & \multicolumn{2}{c}{93.01 ± 0.2} & \multicolumn{2}{c}{88.23 ± 0.3} & \multicolumn{2}{c}{\underline{92.57 ± 0.3}} & \multicolumn{2}{|c}{ 84.63} & \multicolumn{1}{c}{6.5} \\
    \multicolumn{3}{c|}{GCA } & \multicolumn{2}{c}{83.4 ± 0.3} & \multicolumn{2}{c}{72.3 ± 0.1} & \multicolumn{2}{c}{ 80.2 ± 0.4} & \multicolumn{2}{c}{93.10 ± 0.0} & \multicolumn{2}{c}{87.85 ± 0.3} & \multicolumn{2}{c}{92.53 ± 0.2} & \multicolumn{2}{|c}{ 84.89} & \multicolumn{1}{c}{4.0} \\
    \multicolumn{3}{c|}{SimGRACE} & \multicolumn{2}{c}{77.3 ± 0.1} & \multicolumn{2}{c}{71.4 ± 0.1} & \multicolumn{2}{c}{78.3 ± 0.3} & \multicolumn{2}{c}{\underline{93.45 ± 0.4}} & \multicolumn{2}{c}{86.04 ± 0.2} & \multicolumn{2}{c}{91.39 ± 0.4}& \multicolumn{2}{|c}{ 82.98} & \multicolumn{1}{c}{8.5}  \\
    \multicolumn{3}{c|}{COLES } & \multicolumn{2}{c}{81.2 ± 0.4} & \multicolumn{2}{c}{71.5 ± 0.2} & \multicolumn{2}{c}{80.4 ± 0.7} & \multicolumn{2}{c}{92.65 ± 0.1} & \multicolumn{2}{c}{79.64 ± 0.0} & \multicolumn{2}{c}{89.00 ± 0.5} & \multicolumn{2}{|c}{ 82.40} & \multicolumn{1}{c}{ 8.8} \\
    \multicolumn{3}{c|}{ARIEL } & \multicolumn{2}{c}{82.5 ± 0.1} & \multicolumn{2}{c}{72.2 ± 0.2} & \multicolumn{2}{c}{80.5 ± 0.3} & \multicolumn{2}{c}{93.35 ± 0.0} & \multicolumn{2}{c}{88.27 ± 0.2} & \multicolumn{2}{c}{91.43 ± 0.2} & \multicolumn{2}{|c}{ 84.71} & \multicolumn{1}{c}{ 4.8} \\
    \multicolumn{3}{c|}{CCA-SSG } & \multicolumn{2}{c}{\textbf{83.9 ± 0.4}} & \multicolumn{2}{c}{\underline{73.1 ± 0.3}} & \multicolumn{2}{c}{\underline{81.3 ± 0.4}} & \multicolumn{2}{c}{93.37 ± 0.2} & \multicolumn{2}{c}{\underline{88.42 ± 0.3}} & \multicolumn{2}{c}{ 92.44 ± 0.1} & \multicolumn{2}{|c}{ \underline{85.42}} & \multicolumn{1}{c}{ \underline{2.3}} \\
    \midrule
    \multicolumn{3}{c|}{Base Model} & \multicolumn{2}{c}{81.1 ± 0.4} & \multicolumn{2}{c}{71.4 ± 0.1} & \multicolumn{2}{c}{79.1 ± 0.4} & \multicolumn{2}{c}{92.86 ± 0.3} & \multicolumn{2}{c}{87.65 ± 0.2} & \multicolumn{2}{c}{91.19 ± 0.3}&\multicolumn{2}{|c}{83.88}&\multicolumn{1}{c}{9.0} \\
    \multicolumn{3}{c|}{MA-GCL} & \multicolumn{2}{c}{83.3 ± 0.4} & \multicolumn{2}{c}{\textbf{73.6 ± 0.1}} & \multicolumn{2}{c}{\textbf{83.5 ± 0.4}} & \multicolumn{2}{c}{\textbf{94.19 ± 0.1}} & \multicolumn{2}{c}{\textbf{88.83 ± 0.3}} & \multicolumn{2}{c}{\textbf{93.80 ± 0.1}} & \multicolumn{2}{|c}{ \textbf{86.20}} & \multicolumn{1}{c}{ \textbf{1.2}} \\
    \bottomrule[2pt]
    \hspace{0.15cm}
    \end{tabular}%
    
    \caption{Performance of node classification. Here we use public splits on Cora, Citeseer and PubMed.}
      \label{main_table}
\end{table*}

In this section, we conduct experiments on node classification task\footnote{We will explore graph classification and node clustering tasks in Appendix.\textbf{B}. The appendix also includes details of datasets, environment and hyper-parameter settings. The results on hyper-parameter sensitivity and model efficiency are provided as well.} to show the effectiveness of MA-GCL. We start by introducing the experimental setup. Then we present the performance of MA-GCL against SOTA GCL methods on graph benchmarks. We also demonstrate the advantages of each proposed strategy and verify our motivation by extensive experiments.


\subsection{Experimental Setup}
\paragraph{Datasets} We evaluate our approach on six benchmark datasets of node classification, which have been widely used in previous GCL methods. Specifically, citation datasets include Cora, CiteSeer and PubMed {\cite{kipf2016semi}}, co-purchase and co-author datasets include Amazon-Photo, Amazon-Computers and Coauthor-CS {\cite{shchur2018pitfalls}}. 

\paragraph{Evaluation protocol} Following the protocol and implementation of GCA~\cite{zhu2021graph}, we will learn node representations by different GCL methods in an unsupervised manner, and then train the same linear classifier as post-processing for evaluation. We will report classification accuracy as evaluation metric. For three citation datasets, we evaluate the models on the public splits~\cite{kipf2016semi}. For co-purchase and co-author datasets, we randomly split the datasets, where 10\%, 10\%, and the rest 80\% of nodes are selected for the training, validation and test set, respectively {\cite{zhu2021graph,zhang2021canonical}}. For each dataset, we report the average accuracy and standard deviation over $5$ runs in different random seeds. Note that random seeds will also change the splits in co-purchase and co-author datasets.

\begin{table}
	\centering
\vspace{-0.1cm}
	\begin{tabular}{ccc|cccccc}
        \toprule[2pt]
        \multicolumn{3}{c|}{Datasets} & \multicolumn{2}{c}{Cora} & \multicolumn{2}{c}{CiteSeer} & \multicolumn{2}{c}{PubMed}\\
        \midrule
        \multicolumn{3}{c|}{GRACE} & \multicolumn{2}{c}{83.4 ± 1.0} & \multicolumn{2}{c}{69.5 ± 1.4} & \multicolumn{2}{c}{83.1 ± 0.6} \\
        \multicolumn{3}{c|}{GCA} & \multicolumn{2}{c}{82.6 ± 0.9} & \multicolumn{2}{c}{72.2 ± 0.7} & \multicolumn{2}{c}{83.4 ± 0.4} \\
        \multicolumn{3}{c|}{SimGRACE} & \multicolumn{2}{c}{77.2 ± 0.6} & \multicolumn{2}{c}{71.4 ± 0.6} & \multicolumn{2}{c}{81.8 ± 0.7} \\
        \multicolumn{3}{c|}{COLES} & \multicolumn{2}{c}{82.5 ± 0.9} & \multicolumn{2}{c}{72.3 ± 0.6} & \multicolumn{2}{c}{\underline{84.7 ± 0.2}} \\
        \multicolumn{3}{c|}{ARIEL} & \multicolumn{2}{c}{\underline{84.3 ± 1.0}} & \multicolumn{2}{c}{72.7 ± 1.1} & \multicolumn{2}{c}{81.6 ± 0.5} \\
        \multicolumn{3}{c|}{CCA-SSG} & \multicolumn{2}{c}{83.9 ± 0.7} & \multicolumn{2}{c}{\underline{72.7 ± 0.8}} & \multicolumn{2}{c}{84.2 ± 0.2} \\
        \midrule
        \multicolumn{3}{c|}{Base Model} & \multicolumn{2}{c}{82.2 ± 0.7} & \multicolumn{2}{c}{70.7 ± 0.8} & \multicolumn{2}{c}{82.1 ± 0.5} \\
        \multicolumn{3}{c|}{MA-GCL} & \multicolumn{2}{c}{\textbf{84.8 ± 0.4}} & \multicolumn{2}{c}{\textbf{73.3 ± 0.3}} & \multicolumn{2}{c}{\textbf{85.7 ± 0.3}} \\
    \bottomrule[2pt]
    \vspace{-0.2cm}
    \end{tabular}%
    \caption{Performance on Cora, Citeseer and PubMed under random splits.}
    \label{extra exp}
\end{table}
For MA-GCL, we apply our strategies to the base model described in previous section. We use two $h$ operators and multiple $g$ operators in view encoder $f$ for all datasets. We evaluate our model with $K_1=K_2=2$ for Cora and CiteSeer, and $K_1=K_2=1$ for other datasets. More details about hyper-parameter settings are provided in Appendix.\textbf{C}. 

\paragraph{Baselines} We consider a number of node representation learning baselines including very recent SOTA GCL methods. Baselines trained without labels: DGI~\cite{velickovic2019deep}, GRACE~\cite{zhu2020deep}, MVGRL ~\cite{hassani2020contrastive}, BGRL ~\cite{thakoor2021bootstrapped}, GCA ~\cite{zhu2021graph}, COLES ~\cite{zhu2021contrastive}, CCA-SSG ~\cite{zhang2021canonical}, Ariel ~\cite{feng2022adversarial} and SimGRACE ~\cite{xia2022simgrace}. Baselines trained with labels: GCN ~\cite{kipf2016semi}, GAT ~\cite{velickovic2017graph} and InfoGCL ~\cite{xu2021infogcl}. Note that SimGRACE is a graph classification method, we use its variant for node classification tasks by changing GIN~\cite{xu2018powerful} encoders to GCN. All the baselines run with the same evaluation protocol. 

\subsection{Comparison with Baseline Methods}
We report the performance of node classification in Table ~\ref{main_table}. We \textbf{bold} the best method trained
 without labels in each column, and \underline{underline} the best performed baseline. We can see that MA-GCL can achieve SOTA performance on 5 out of 6 graph benchmarks, and the relative improvement can go up to $2.7\%$. Considering that the public splits on Cora, Citeseer and PubMed might not be representative, we also investigate another benchmark setting~\cite{feng2022adversarial} with random splits on these three datasets, and compare with the most competitive baselines. As shown in Table~\ref{extra exp}, MA-GCL consistently outperforms baseline methods, which demonstrates the effectiveness of MA-GCL.
\begin{table*}[htbp]
  \centering
  
  \footnotesize

    \begin{tabular}{ccc|cccccccccccc|ccc}
    \toprule[2pt]
    \multicolumn{3}{c|}{Datasets} & \multicolumn{2}{c}{Cora} & \multicolumn{2}{c}{CiteSeer} & \multicolumn{2}{c}{PubMed} & \multicolumn{2}{c}{Coauthor-CS} & \multicolumn{2}{c}{Amazon-C} & \multicolumn{2}{c}{Amazon-P}& \multicolumn{2}{|c}{Avg. Acc.}& \multicolumn{1}{c}{Avg. Rank} \\
    \midrule
    \multicolumn{3}{c|}{Base Model+A} & \multicolumn{2}{c}{82.5 ± 0.5} & \multicolumn{2}{c}{73.1 ± 0.4} & \multicolumn{2}{c}{82.3 ± 0.3} & \multicolumn{2}{c}{93.41 ± 0.0} & \multicolumn{2}{c}{88.13 ± 0.2} & \multicolumn{2}{c}{93.10 ± 0.1}& \multicolumn{2}{|c}{85.42}& \multicolumn{1}{c}{4.2} \\
    \multicolumn{3}{c|}{Base Model+R} & \multicolumn{2}{c}{82.4 ± 0.3} & \multicolumn{2}{c}{72.5 ± 0.1} & \multicolumn{2}{c}{82.8 ± 0.4} & \multicolumn{2}{c}{93.12 ± 0.1} & \multicolumn{2}{c}{88.06 ± 0.1} & \multicolumn{2}{c}{92.26 ± 0.3} & \multicolumn{2}{|c}{85.19}& \multicolumn{1}{c}{6.3}\\
    \multicolumn{3}{c|}{Base Model+S} & \multicolumn{2}{c}{81.9 ± 0.2} & \multicolumn{2}{c}{72.7 ± 0.2} & \multicolumn{2}{c}{81.8 ± 0.1} & \multicolumn{2}{c}{93.13 ± 0.0} & \multicolumn{2}{c}{87.94 ± 0.2} & \multicolumn{2}{c}{92.66 ± 0.2}& \multicolumn{2}{|c}{85.02}& \multicolumn{1}{c}{6.5} \\
    
    \midrule
    \multicolumn{3}{c|}{MA-GCL w/o A} & \multicolumn{2}{c}{82.5 ± 0.6} & \multicolumn{2}{c}{73.3 ± 0.4} & \multicolumn{2}{c}{82.9 ± 0.4} & \multicolumn{2}{c}{93.31 ± 0.1} & \multicolumn{2}{c}{88.09 ± 0.2} & \multicolumn{2}{c}{92.86 ± 0.3}& \multicolumn{2}{|c}{85.49}& \multicolumn{1}{c}{4.3} \\
    \multicolumn{3}{c|}{MA-GCL w/o R} & \multicolumn{2}{c}{83.3 ± 0.6} & \multicolumn{2}{c}{73.0 ± 0.0} & \multicolumn{2}{c}{83.1 ± 0.4} & \multicolumn{2}{c}{93.47 ± 0.0} & \multicolumn{2}{c}{88.67 ± 0.2} & \multicolumn{2}{c}{93.41 ± 0.2} & \multicolumn{2}{|c}{85.83}& \multicolumn{1}{c}{2.8}\\
    \multicolumn{3}{c|}{MA-GCL w/o S} & \multicolumn{2}{c}{82.7 ± 0.3} & \multicolumn{2}{c}{73.1 ± 0.4} & \multicolumn{2}{c}{83.3 ± 0.5} & \multicolumn{2}{c}{93.92 ± 0.1} & \multicolumn{2}{c}{88.34 ± 0.4} & \multicolumn{2}{c}{93.01 ± 0.0}& \multicolumn{2}{|c}{85.73}& \multicolumn{1}{c}{2.8} \\
    
    \midrule
    \multicolumn{3}{c|}{Base Model} & \multicolumn{2}{c}{81.1 ± 0.4} & \multicolumn{2}{c}{71.4 ± 0.1} & \multicolumn{2}{c}{79.1 ± 0.4} & \multicolumn{2}{c}{92.86 ± 0.3} & \multicolumn{2}{c}{87.65 ± 0.2} & \multicolumn{2}{c}{91.19 ± 0.3} &\multicolumn{2}{|c}{83.88}&\multicolumn{1}{c}{8.0}\\
    \midrule
    \multicolumn{3}{c|}{MA-GCL} & \multicolumn{2}{c}{83.3 ± 0.4} & \multicolumn{2}{c}{73.6 ± 0.1} & \multicolumn{2}{c}{83.5 ± 0.4} & \multicolumn{2}{c}{94.19 ± 0.1} & \multicolumn{2}{c}{88.83 ± 0.3} & \multicolumn{2}{c}{93.80 ± 0.1}  & \multicolumn{2}{|c}{ {86.20}} & \multicolumn{1}{c}{ {1.0}}\\
    \bottomrule[2pt]
    \end{tabular}%
    \vspace{0.18cm}
    \caption{Ablation Studies of MA-GCL. Base Model ($0$ strategy), Base Model+X ($1$ strategy), MA-GCL w/o X ($2$ strategies), and MA-GCL ($3$ strategies).}
     \label{ablation_table}
\end{table*}%
\subsection{Ablation Studies}

We conduct ablation experiments to prove the effectiveness of each model augmentation strategy: \textbf{A}symmetric, \textbf{R}andom and \textbf{S}huffling. We compare the full model ($3$ strategies) with base model ($0$ strategy) and six ablated models ($1$ or $2$ strategies). A model with or without asymmetric strategy determines whether two view encoders have the same number of $g$ operators; a model with or without random strategy determines whether the encoder architectures are fixed or randomized in every epoch; a model with or without shuffling strategy determines whether two view encoders have different operator permutations. Note that the asymmetric strategy indicates that the permutations of two view encoders cannot be exactly the same. Thus for models with asymmetric strategy but without shuffling strategy, we will force $K_i=K_i'$ for $i<N$. The results are reported in Table~\ref{ablation_table}.




In summary, if an ablated model is further combined with asymmetric/random/shuffling strategy, its accuracy can be respectively improved by 0.86/0.65/0.54\% on average. Hence all three strategies have positive effects on the overall performance of MA-GCL, and the asymmetric strategy is the most effective one among the three. Also note that the performance of our base model is weaker than that of SOTA GCL baselines, which validates the superiority of our paradigm of model augmentation.



\subsection{Motivation Verification} 
\label{sec:mutual}
\begin{figure}[] 
\centering
    \subfigure[CiteSeer]{
        \label{MI1}
        \includegraphics[scale=0.16]{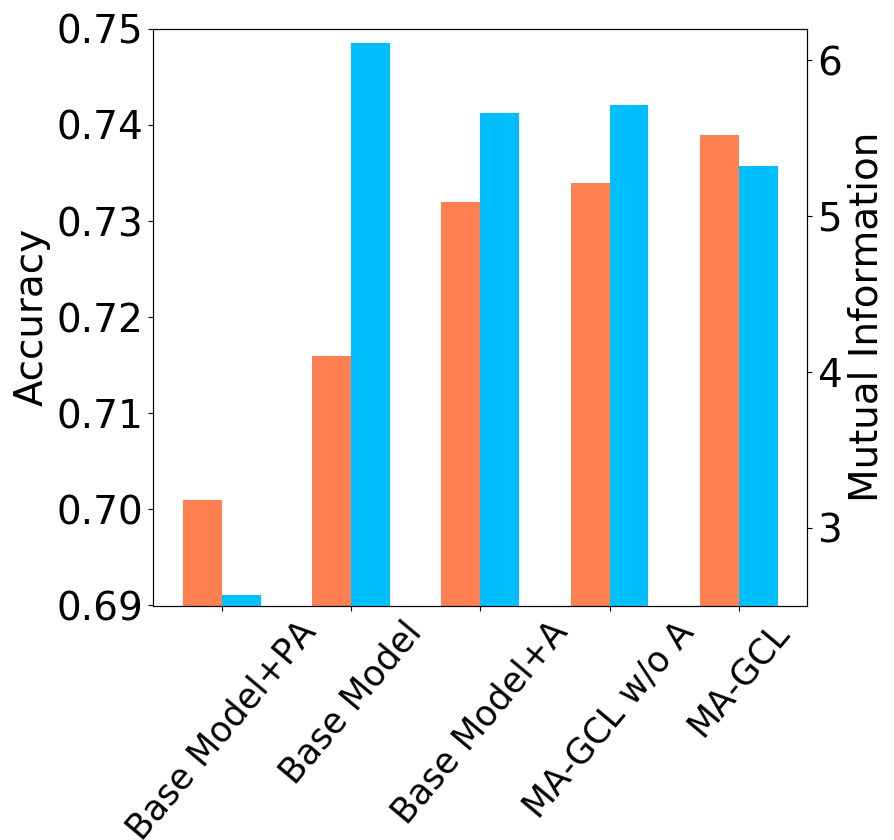}
    }
    \hspace{-1.1cm}
    \hfill
    \subfigure[Amazon-Photo]{
        \label{MI2}
        \includegraphics[scale=0.17]{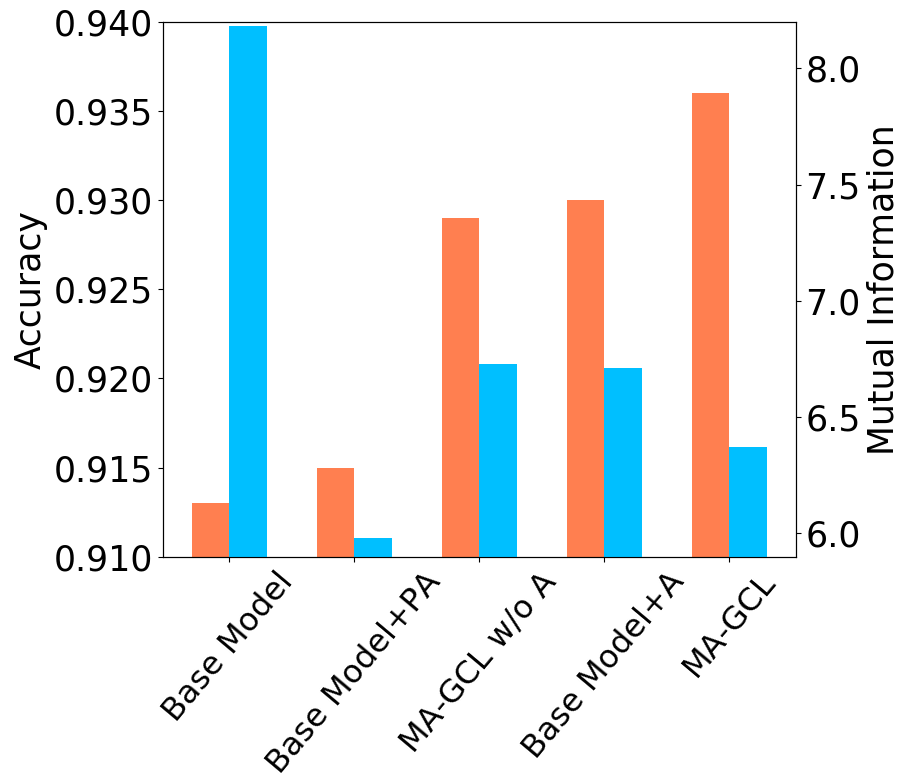}
    }
\caption{Experimental results for motivation verification. Orange columns (estimation of area D in Fig.~\ref{fig:vn}) denote the classification accuracies of different methods; Blue columns (estimation of area C+D in Fig.~\ref{fig:vn}) denote the mutual information between two views. A method with higher orange column and lower blue column indicates that more task-relevant knowledge can be encoded via CL with less irrelevant noises. The other four datasets are in Appendix.\textbf{B}.}
\label{fig:mutual}
\end{figure}

Now we will conduct experiments to verify our motivation that (1) typical GDA techniques alone (\textit{e.g.}, edge dropping) cannot generate diverse enough augmentations, and thus will bring much noise (area C in Fig.~\ref{fig:vn}) into learned representations; (2) directly perturbing model parameters may harm the semantics (area D in Fig.~\ref{fig:vn}) in encoded representations. 

\paragraph{Setup} We propose to estimate the information in area C and D for different GCL methods. According to the InfoMin principle~\cite{tian2020makes}, a method is better if it has a larger area D (\textit{i.e.,} task-relevant information) and a smaller area C (\textit{i.e.,} task-irrelevant noises). However, it is hard to directly estimate the information in area C or D alone. Thus we employ MINE~\cite{belghazi2018mutual} to compute the mutual information between two views, as an estimation of \textbf{area C+D}. Given the representation trained by CL, we will use its accuracy on downstream tasks as an estimation of \textbf{area D}. For a fair comparison, we investigate five methods based on our backbone model: \textbf{Base Model} represents typical GCL methods with random data augmentations; \textbf{Base Model+PA} adds parameter perturbations to the base model, which represents a recent paradigm~\cite{xia2022simgrace} of parameter augmentation (PA); \textbf{Base Model+A}, \textbf{MA-GCL w/o A} and \textbf{MA-GCL} are employed to demonstrate our effectiveness. 

\paragraph{Results} As shown in Fig.~\ref{fig:mutual}, we have the following observations: (1) Base Model always has the highest mutual information but the worst two accuracies, which supports our statement that views in typical GCL methods are too close to each other and bring much noises; (2) Base Model+PA has low accuracies and mutual information, which shows that perturbing model parameters will significantly harm the semantics in encoded representations; (3) By comparing Base Model against Base Model+A, and MA-GCL w/o A against MA-GCL, we can see that the asymmetric strategy can effectively push two views away from each other, result in a significant decrease of blue columns, and thereby reduce task-irrelevant noises. The above observations validate our motivation and strategy design.





\section{Conclusion}
\label{conclusion}
In this paper, we highlight a key limitation of previous GCL methods that their contrastive views are too close to effectively filter out noises. Then we propose a novel paradigm named model augmentation, which focuses on manipulating the neural architectures of GNN view encoders instead of perturbing graph inputs or model parameters. Specifically, we present three model augmentation tricks for GCL: \textit{asymmetric}, \textit{random} and \textit{shuffling}, which can respectively help alleviate high-frequency noises, enrich training instances and bring safer augmentations. With the three tricks, two contrastive views can keep a proper distance from each other, \textit{i.e.}, neither too far (losing task-relevant semantics) nor too close (introducing unnecessary noises). Experiments show that MA-GCL can achieve SOTA performance by applying the three tricks on a simple base model as a plug-and-play component. We hope this work can 
provide a new direction of GCL, and future work may consider generalize the idea of model augmentation to graphs with heterophily.

\section{Acknowledgments.}
This work is supported in part by the National Natural Science Foundation of China (No. U20B2045, 62192784, 62172052, 62002029, 62006129, 62172052, U1936014).
\bibliography{aaai2023}

\end{document}


\appendix
\section{Proof of Theorem 4.1}
\label{sec:proof}

\textit{Proof }: Since $\bm{U}$ is invertible, we can alternatively optimize $\bm{P}=\bm{U}^T\bm{W}$ instead of $\bm{W}$ in the contrastive loss. Then the optimization in Eq.~($4$) can be reformalized as 
\begin{equation}
    \min_{\bm{P}} \mbox{tr}(\bm{M}\bm{M}^T), \quad \mbox{subject to}\quad  \bm{P}^T\bm{P}=\bm{I},
\label{eq:asym2}
\end{equation}
where $\bm{M}=\bm{U}(\Lambda^{L}-\Lambda^{L'})\bm{P}$. To solve the optimization problem, we first relax the condition to $\bm{P}_i^T\bm{P}_i=1$ for $i=1,2\dots d_O$, where $\bm{P}_i$ is the $i$-th column of $\bm{P}$. Then the Lagrangian function is
\begin{equation}
\begin{aligned}
\mathcal{L}(\bm{P},\beta)&=\mbox{tr}(\bm{U}(\Lambda^{L}-\Lambda^{L'})\bm{P}\bm{P}^T(\Lambda^{L}-\Lambda^{L'})\bm{U}^T)-\sum_{i=1}^{d_O}\beta_i (\bm{P}_i^T\bm{P}_i-1)\\
    &=\mbox{tr}((\Lambda^{L}-\Lambda^{L'})^2 \bm{P}\bm{P}^T)-\sum_{i=1}^{d_O}\beta_i (\bm{P}_i^T\bm{P}_i-1),
\end{aligned}
\end{equation}
where $\beta=(\beta_1,\beta_2\dots \beta_{d_O})$ is the Lagrangian multiplier.

With the Karush$-$Kuhn$-$Tucker (KKT) conditions, we have
\begin{equation}
\begin{aligned}
\frac{\partial \mathcal{L}(\bm{P},\beta)}{\partial \bm{P}}&=0\\
2(\Lambda^{L}-\Lambda^{L'})^2\bm{P}-2\bm{P}\mbox{diag}(\beta)&=0\\
(\Lambda^{L}-\Lambda^{L'})^2\bm{P}&
= \bm{P}\mbox{diag}(\beta).
\end{aligned}
\end{equation}
Therefore, the columns of $\bm{P}$ are the eigenvectors of matrix $(\Lambda^{L}-\Lambda^{L'})^2$. Note that the original condition $\bm{P}^T\bm{P}=\bm{I}$ requires that the columns of $\bm{P}$ are orthogonal. Hence the optimal $\bm{P}$ is the eigenvectors of $(\Lambda^{L}-\Lambda^{L'})^2$ with minimum eigenvalues:
\begin{equation}
\bm{P}^*=\bm{I}[k_1,k_2,\dots k_{d_O}],
\end{equation}
where $1\leq k_1<k_2<\dots k_{d_O}\leq |\mathcal{V}|$ are the best $k$s that minimize $(\lambda_k^L-\lambda_k^{L'})^2$, and $\bm{I}[k_1,k_2,\dots k_{d_O}]$ denotes the corresponding columns of the identity matrix $\bm{I}$. QED.

\section{Additional Experiments}
\label{sec:add}
\subsection{Graph Classification Experiments}
\label{sec:graph}
\begin{figure}[] 
\centering
    \subfigure[NCI1]{
        \label{NCI1}
        \includegraphics[scale=0.25]{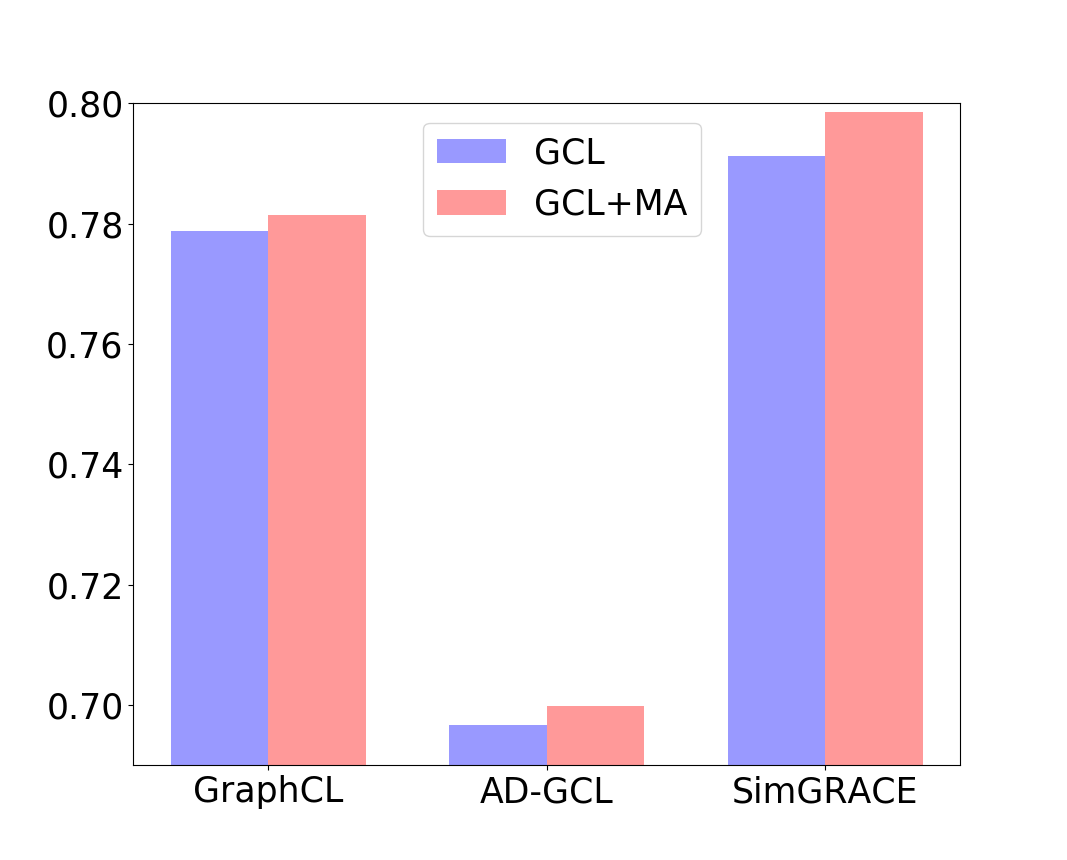}
    }
    \hspace{-1.1cm}
    \hfill
    \subfigure[MUTAG]{
        \includegraphics[scale=0.25]{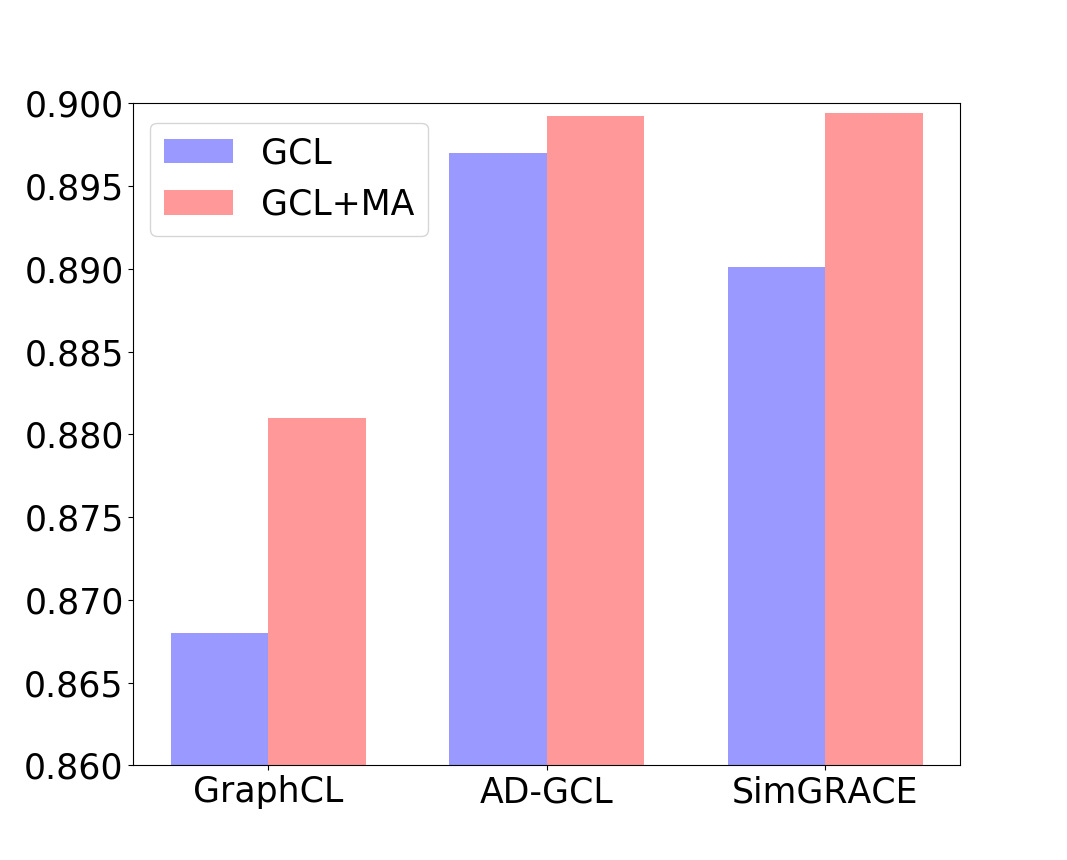}
        \label{MUTAG}
    }
    \subfigure[COLLAB]{
        \label{COLLAB}
        \includegraphics[scale=0.25]{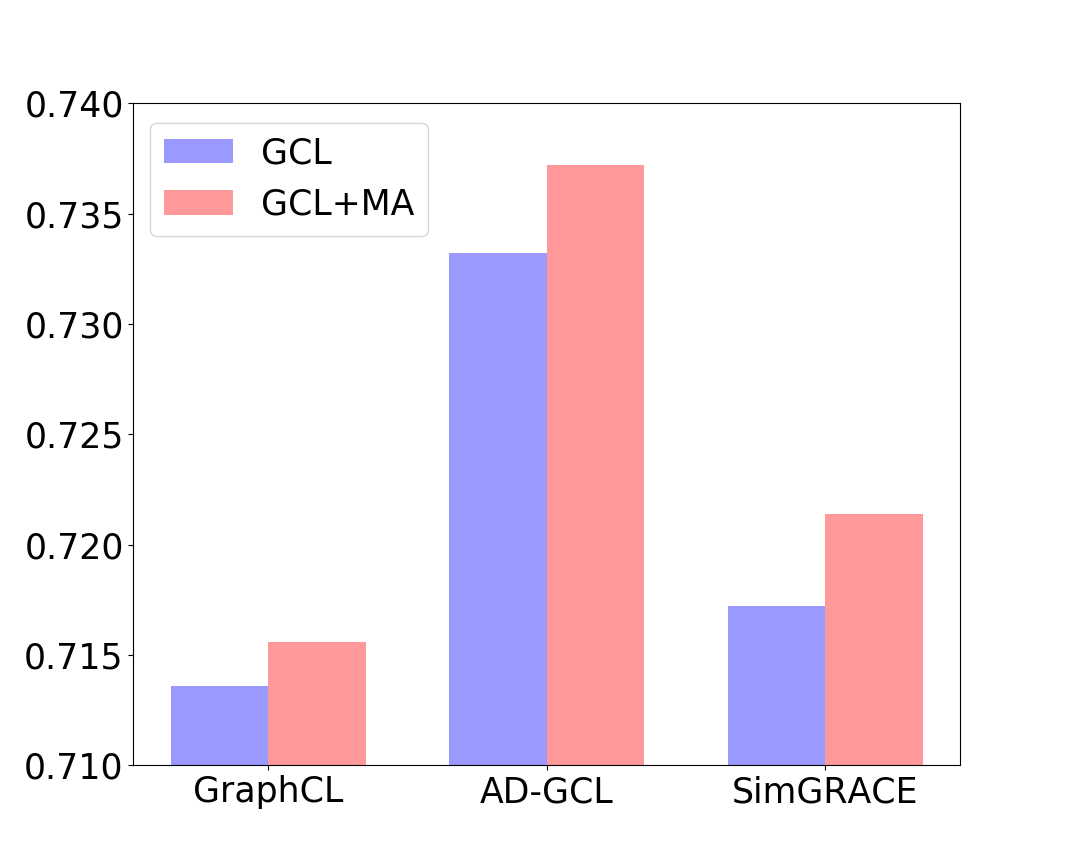}
    }
    \hspace{-1.1cm}
    \hfill
    \subfigure[IMDB-B]{
        \includegraphics[scale=0.25]{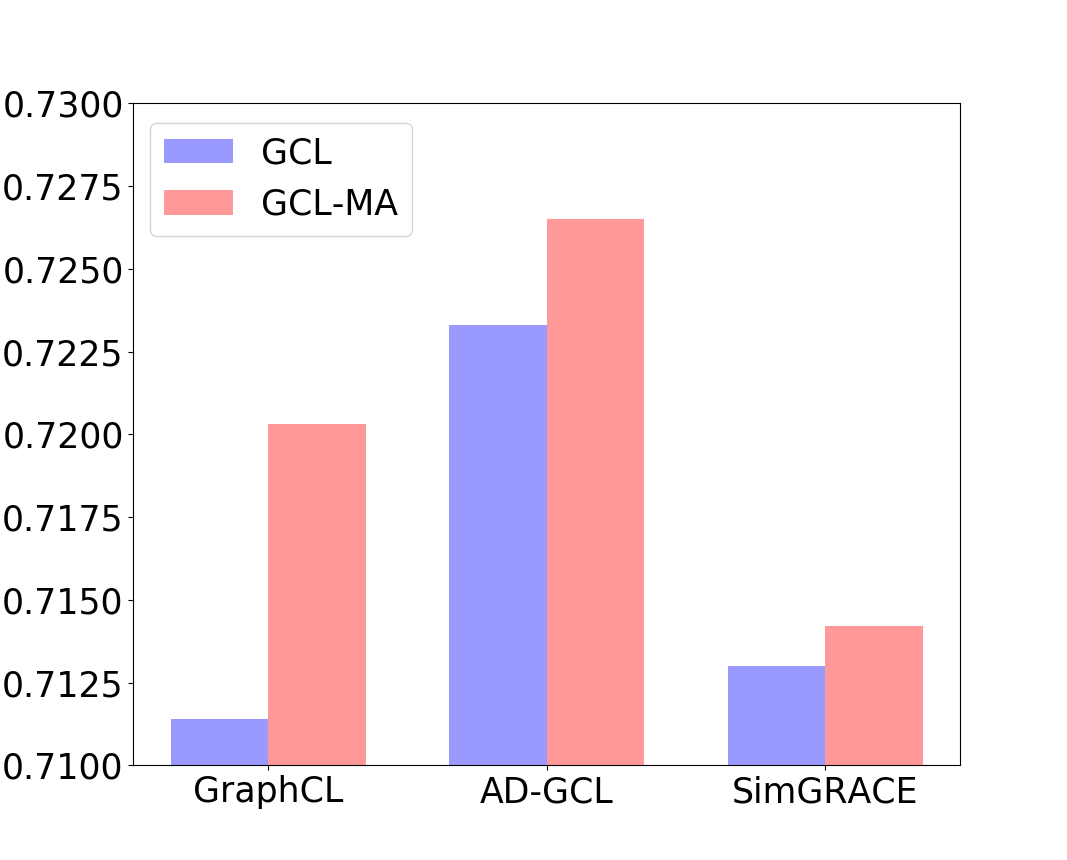}
        \label{IMDB-B}
    }
    \caption{Experimental results of graph classification accuracy on four datsets. Every pair of columns represent an original GCL model and its variant with our model augmentation mechanism, respectively.}
\label{fig:gc1}
\end{figure}

\begin{table}[htbp]
  \centering
  \caption{Datasets statistics for graph classification.}
    \begin{tabular}{cccccccccc}
    \toprule[2pt]
    \multicolumn{2}{c}{Dataset} & \multicolumn{2}{c}{\#Graphs} & \multicolumn{2}{c}{Avg. \#Nodes} & \multicolumn{2}{c}{Avg. \#Edges} & \multicolumn{2}{c}{\#Classes} \\
    \midrule
    \multicolumn{2}{c}{NCI1} & \multicolumn{2}{c}{4,110} & \multicolumn{2}{c}{29.87} & \multicolumn{2}{c}{32.3} & \multicolumn{2}{c}{2} \\
    \multicolumn{2}{c}{MUTAG} & \multicolumn{2}{c}{188} & \multicolumn{2}{c}{17.93} & \multicolumn{2}{c}{19.79} & \multicolumn{2}{c}{2} \\
    \multicolumn{2}{c}{COLLAB} & \multicolumn{2}{c}{5,000} & \multicolumn{2}{c}{74.5} & \multicolumn{2}{c}{2457.78} & \multicolumn{2}{c}{3} \\
    \multicolumn{2}{c}{IMDB-BINARY} & \multicolumn{2}{c}{1,000} & \multicolumn{2}{c}{13} & \multicolumn{2}{c}{65.94} & \multicolumn{2}{c}{2} \\
    \bottomrule[2pt]
    \end{tabular}%
  \label{gc_data}%
\end{table}%
Though our strategies are mostly derived from node-level representation learning, we conduct experiments to further validate our idea of model augmentation on $4$ graph classification benchmarks: NCI1, MUTAG, COLLAB and IMDB-BINARY in TUDatasets~\cite{morris2020tudataset}. The statistics of datasets can be found in Table \ref{gc_data}. Specifically, we apply the three strategies to three state-of-the-art GCL models for graph-level representation learning, including GraphCL~\cite{you2020graph}, AD-GCL~\cite{suresh2021adversarial} and SimGRACE~\cite{xia2022simgrace}. Then we have their model augmented versions as GraphCL+MA, AD-GCL+MA and SimGRACE+MA, and compare the classification performance with their original models. We fix the range of $K$ as $[0, 2]$ and the range of $K'$ as $[0, 4]$. For AD-GCL+MA, we remove the component of adaptive augmentations in AD-GCL, and only employ the backbone with feature and edge dropping. We use GIN~\cite{xu2018powerful} as the encoder for all methods and all other experimental settings are the same as original models~\cite{you2020graph, suresh2021adversarial, xia2022simgrace}. We report the results in Fig.~\ref{fig:gc1}. As we can see in the results, our model augmentation mechanism can be successfully applied to the SOTA GCL models for graph-level modeling, and achieve better performance on downstream graph classification tasks. 

\subsection{Motivation Verification Experiments}
\label{mutual information exp extral}
\begin{figure}[] 

\centering
    \subfigure[Cora]{
        \label{MI4}
        \includegraphics[scale=0.19]{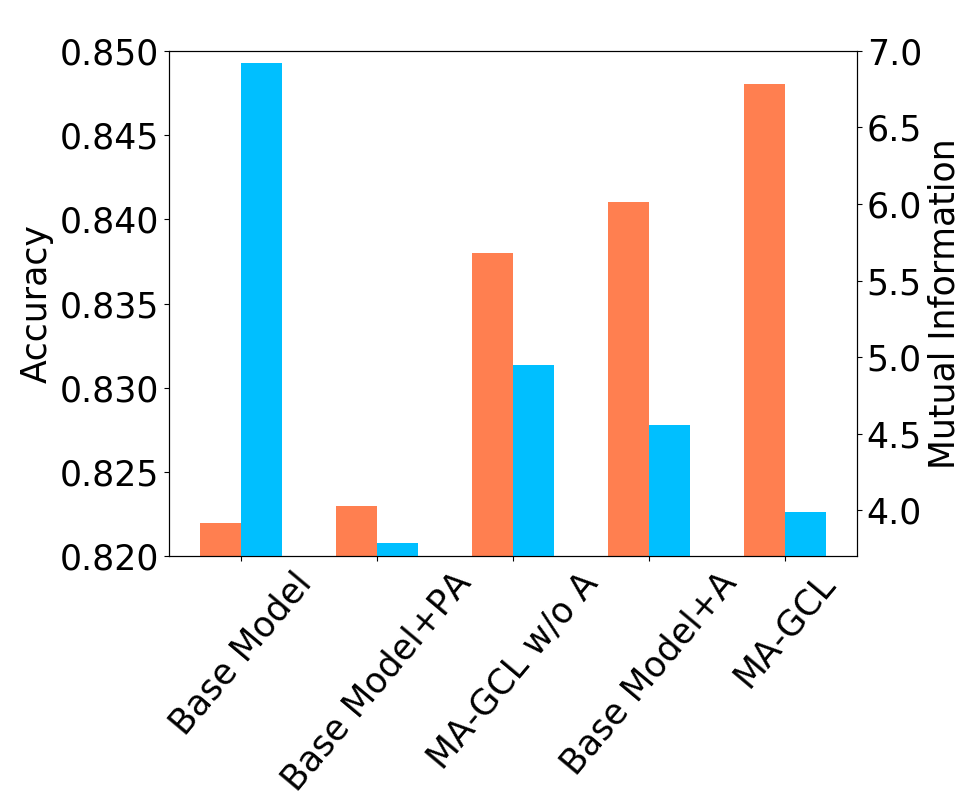}
    }
    \hspace{-1.1cm}
    \hfill
    \subfigure[PubMed]{
        \includegraphics[scale=0.19]{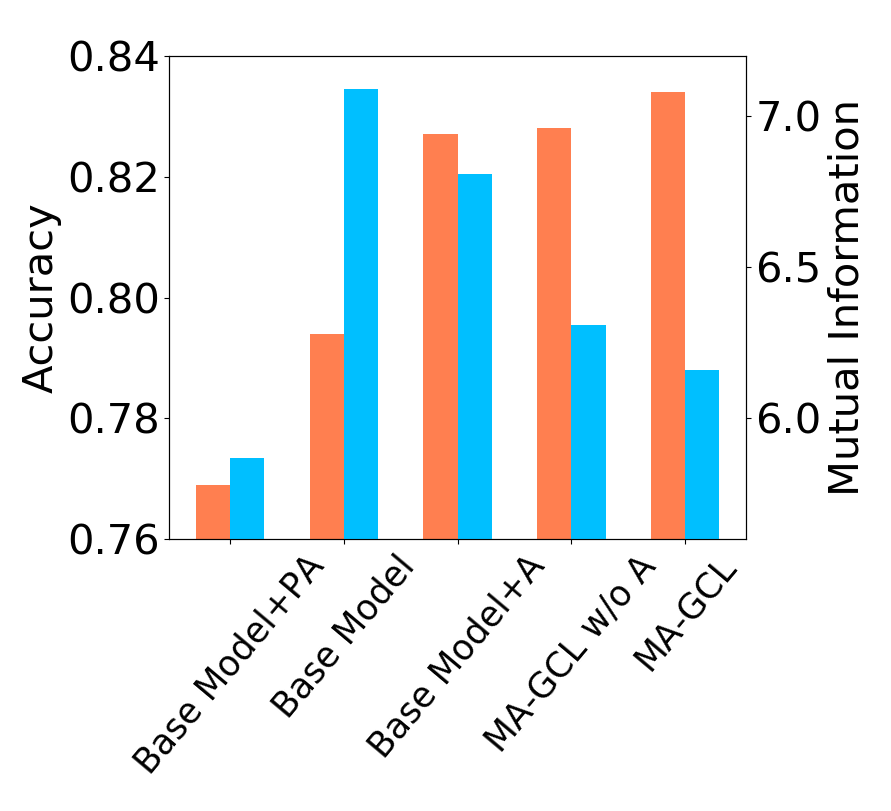}
        \label{MI5}
    }
    \hspace{-1.1cm}
    \hfill
    \subfigure[Amazon-C]{
        \includegraphics[scale=0.19]{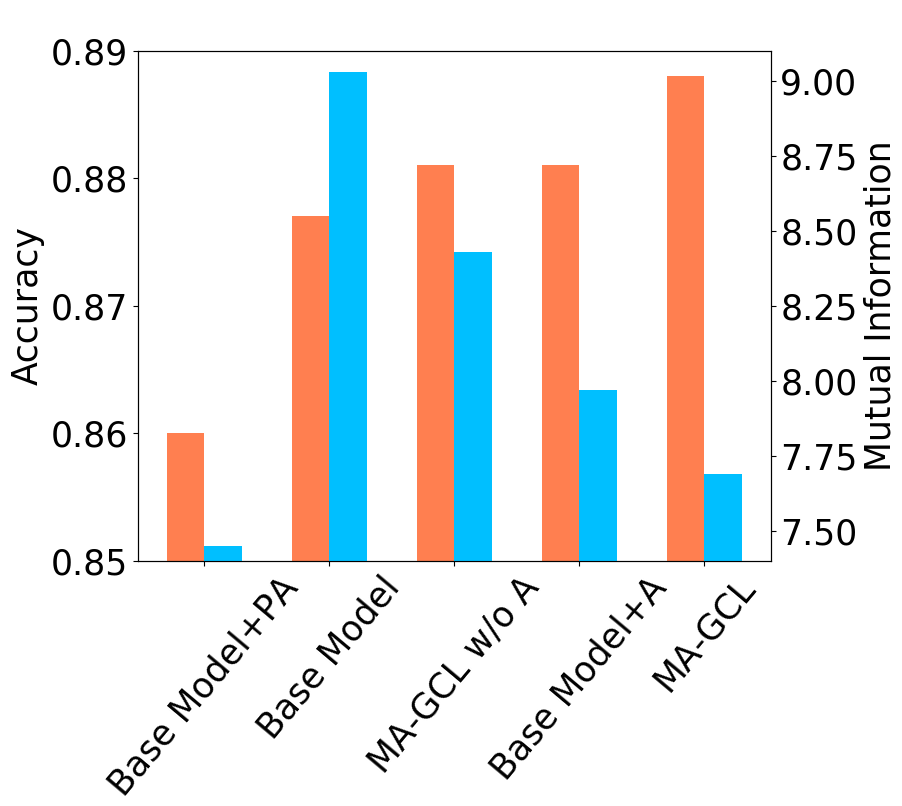}
        \label{MI6}
    }
    \caption{Experimental results of motivation verification on Cora, Amazon-Computer and PubMed.}
\label{fig:mutual2}
\end{figure}

We present motivation verification experiments on Cora, PubMed and Amazon-C in Fig.~\ref{fig:mutual2}. All the settings are the same as in Section 5.4. The patterns on Cora, PubMed, Amazon-C are also consistent with those in Section 5.4.

\subsection{Hyper-parameter Experiments}
\label{hyper exp}
\begin{figure}[] 
\centering
    \subfigure[Cora]{
        \label{Heat:cora}
        \includegraphics[scale=0.30]{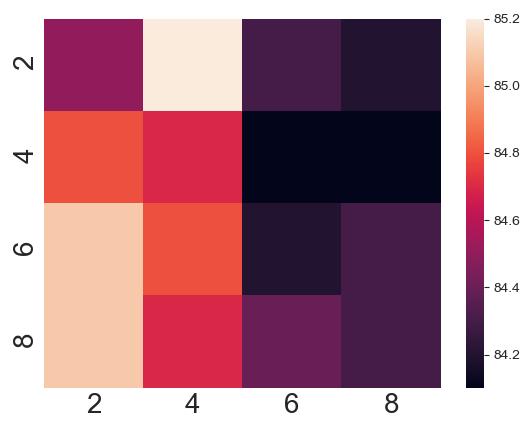}
    }
    \hspace{-1.1cm}
    \hfill
    \subfigure[CiteSeer]{
        \includegraphics[scale=0.30]{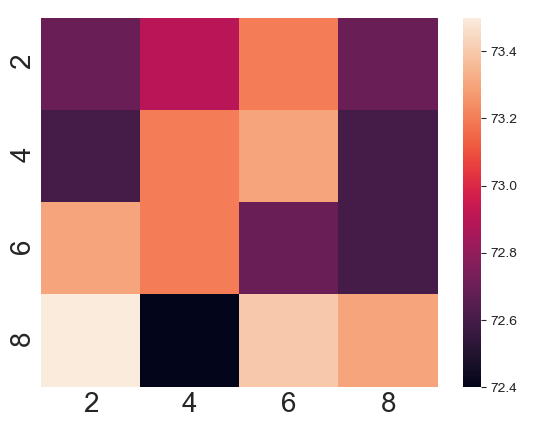}
        \label{Heat:citeseer}
    }
    \hspace{-1.1cm}
    \hfill
    \subfigure[PubMed]{
        \includegraphics[scale=0.30]{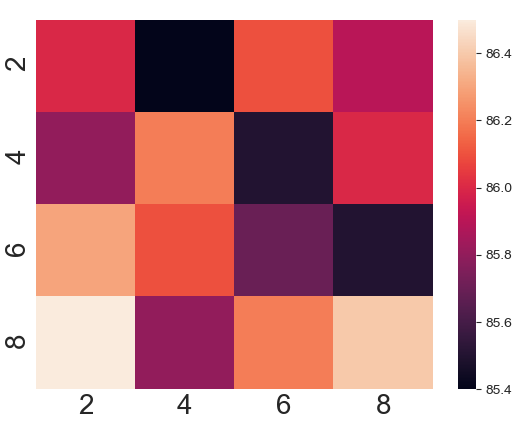}
        \label{heat:pubmed}
    }
    \caption{Visualizations of the accuracies under different range combinations of $K\in [0,high]$ and $K'\in [0,high']$. Each row represents a specific setting for $high'$ and each column represents a specific setting for $high$.}
    \label{heatmap}
\end{figure}
\paragraph{Effect of Random Ranges} We explore the influence of the ranges of $K$ and $K'$ on node classification. Here we try different range combinations. Specifically, we fix the $low$ and $low'$ as $0$, and report the corresponding performance under different combinations of $high$ and $high'$ on Cora, CiteSeer and PubMed (random splits). Here the performance is the average accuracy over 5 runs in different random seeds. The experimental results are shown in Fig. \ref{heatmap}. We can find that (1) By simply setting the range of $K$ as $[0, 2]$ and the range of $K'$ as $[0, 8]$, MA-GCL can achieve SOTA performance on all the three benchmarks. The results are even better than we reported in the main document. According to our theory of the asymmetric strategy, a wider gap between $K$ and $K'$ can help better alleviate high-frequency noises. Also note that negative samples will adopt the view encoder corresponding to $K$ instead of $K'$. Hence for the range pair of $K$ and $K'$, $[2,8]$ is much better than $[8,2]$ due to the over-smoothing issue. (2) MA-GCL is not very sensitive to the ranges of $K$ and $K'$. The difference between the best accuracy and the worst one is less than one percent on three benchmarks. 

\begin{figure}[] 

\centering
\includegraphics[scale=0.3]{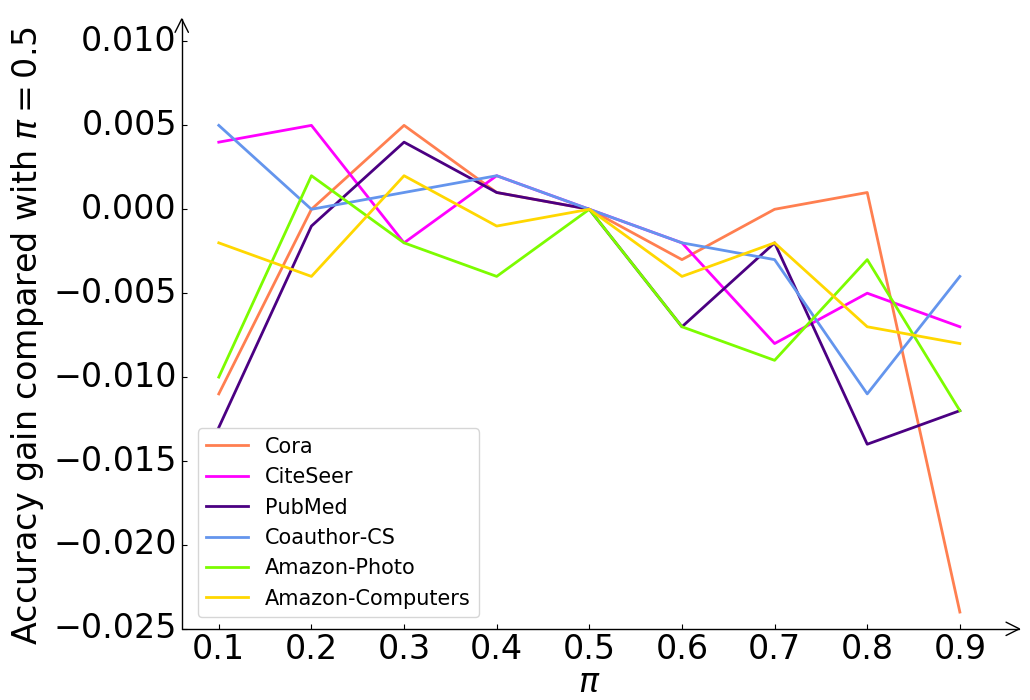}
\hspace{-0.4cm}
\caption{Comparisons among graph filters with different $\pi$. Lines with different colors represent the results on different datasets. For convenience, we subtract the accuracy of $\pi=0.5$ for each line.}
\label{fig:pi}
\end{figure}

\paragraph{Effect of Graph Filters}
\label{filter}
We conduct experiments to explore the effects of graph filter $\bm{F}$ in MA-GCL. Note that we use fixed $\bm{F}={(1-\pi)}\bm{I}+\pi \bm{D}^{-\frac{1}{2}}\bm{A}\bm{D}^{-\frac{1}{2}}$ by setting $\pi=0.5$ in previous sections. In this section, we try different graph filters by choosing $\pi \in (0,1)$, and report the results on the $6$ benchmarks. We randomly split the datasets, where $10\%$, $10\%$, and the rest $80\%$ of nodes are selected for training, validation and test sets. We report the average accuracy over 5 runs in different random seeds. Other hyper-parameters are the same as reported in Table ~\ref{tab:hyper1} and ~\ref{tab:hyper2}. Experimental results are shown in Fig. \ref{fig:pi}. We find that the optimal $\pi$ of every dataset is less than $0.5$. Recall that the eigenvalues of $\bm{F}$ approximately fall in range $[1-\frac{3}{2}\pi, 1]$, and the function $\min_\lambda(\lambda^L-\lambda^{L'})^2$ prefers $\lambda\rightarrow 0$ or $\lambda\rightarrow 1$. Hence a smaller $\pi$ can help GCL preserve the information corresponding to $\lambda\rightarrow 1$, and filter out high-frequency noise (\textit{i.e.}, the graph signals corresponding to $\lambda\rightarrow 0$). This observation can also validate our theory of the asymmetric strategy. 

\paragraph{Effect of Hidden Size of Encoder and Projector} 
\begin{figure}[] 
\centering
    \subfigure[Encoder]{
        \label{encoder_size}
        \includegraphics[scale=0.25]{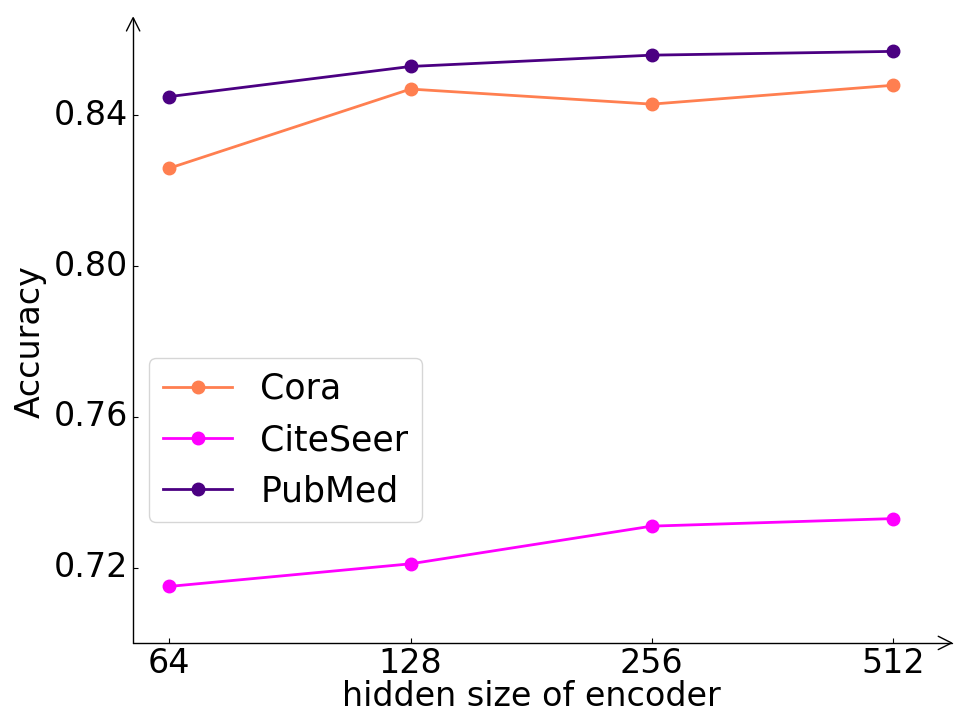}
    }
    \hspace{-1.1cm}
    \hfill
    \subfigure[Projector]{
        \includegraphics[scale=0.25]{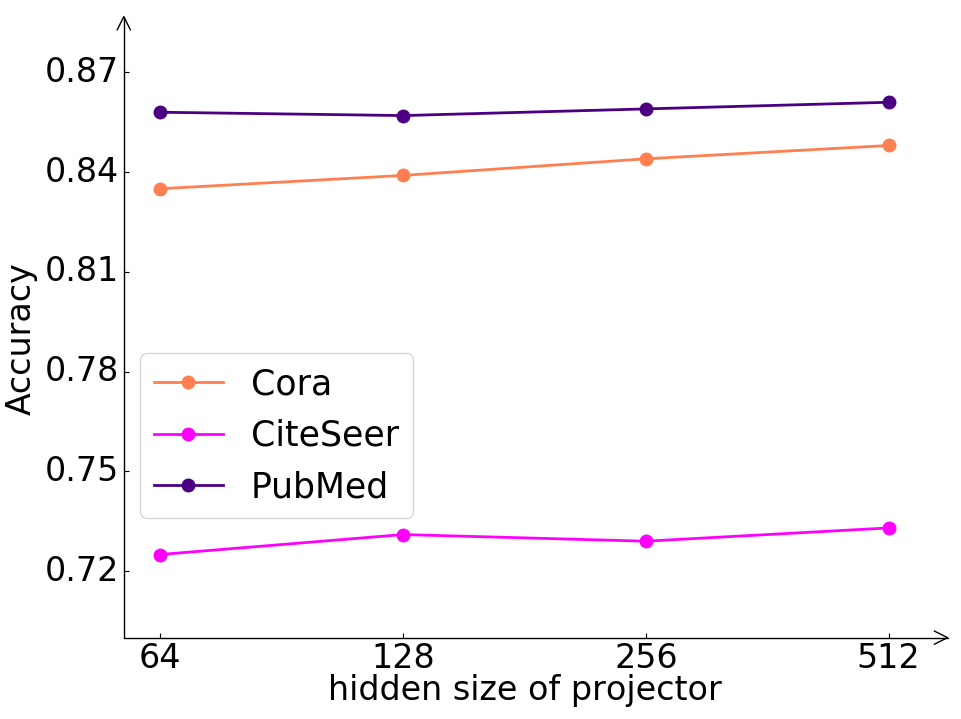}
        \label{proj_size}
    }
    \caption{Visualization of the trend of accuracy with respect to hidden size of encoder and projector. Lines with different colors represent the results on different datasets. }
    \label{hidden}
\end{figure}
We explore the effects of the hidden sizes of graph encoder and embedding projector on node classification tasks. Here the experiments are conducted on three citation networks of Cora, CiteSeer and PubMed. Specifically, we fix all other hyper-parameters and change the hidden size of encoder within the value of $\{64, 128, 256, 512\}$ and report the trend of performance in Fig.~\ref{encoder_size}. Similarly, we also conduct the same experiments with the hidden size of embedding projector, and report the results in Fig.~\ref{proj_size}. We can see that the overall performance is stable with respect to hidden sizes. As the hidden sizes increase, the performance on node classification increases smoothly.

\section{More Detailed Experimental Settings}
\label{sec:detail}

\subsection{Datasets}
We evaluate our models on six node classification benchmarks include Cora, CiteSeer, PubMed, Coauthor-CS,  Amazon-Computer and Amazon-Photo. The statistics of datasets is shown in Table \ref{tab:data}. 
\begin{table}[htbp]
  \centering
  
  \caption{Statistics of datasets.}
    \begin{tabular}{cccccccccc}
    \toprule[2pt]
    \multicolumn{2}{c}{Dataset} & \multicolumn{2}{c}{\#Nodes} & \multicolumn{2}{c}{\#Edges} & \multicolumn{2}{c}{\#Features} & \multicolumn{2}{c}{\#Classes} \\
    \midrule
    \multicolumn{2}{c}{Cora} & \multicolumn{2}{c}{2,708} & \multicolumn{2}{c}{10,556} & \multicolumn{2}{c}{133} & \multicolumn{2}{c}{7} \\
    \multicolumn{2}{c}{CiteSeer} & \multicolumn{2}{c}{3,327} & \multicolumn{2}{c}{9,228} & \multicolumn{2}{c}{3,703} & \multicolumn{2}{c}{6} \\
    \multicolumn{2}{c}{PubMed} & \multicolumn{2}{c}{19,717} & \multicolumn{2}{c}{88,651} & \multicolumn{2}{c}{500} & \multicolumn{2}{c}{3} \\
    \multicolumn{2}{c}{Coauthor-CS} & \multicolumn{2}{c}{18,333} & \multicolumn{2}{c}{327,576} & \multicolumn{2}{c}{6,805} & \multicolumn{2}{c}{15} \\
    \multicolumn{2}{c}{Amazon-P} & \multicolumn{2}{c}{7,650} & \multicolumn{2}{c}{287,326} & \multicolumn{2}{c}{745} & \multicolumn{2}{c}{8} \\
    \multicolumn{2}{c}{Amazon-C} & \multicolumn{2}{c}{13,752} & \multicolumn{2}{c}{574,418} & \multicolumn{2}{c}{767} & \multicolumn{2}{c}{10} \\
    \bottomrule[2pt]
    \end{tabular}%
    \label{tab:data}
\end{table}%
\paragraph{Cora, CiteSeer, PubMed} are three commonly used node classification datasets, each of which contains one citation network, where nodes mean papers and edges represent citation relationships.

\paragraph{Coauthor-CS} is a widely used node classification benchmarks based on the Microsoft Academic Graph from the KDD Cup 2016 challenge~\cite{sinha2015overview}. The vertices represent authors and edges mean the co-author relationship between authors. The ground-truth labels of authors indicate most active fields of study for authors, and the keywords of the author's papers are extracted as the features of the vertices. 

\paragraph{Amazon-Photo, Amazon-Computers} are node classification datasets based on the Amazon co-purchase graph. The nodes mean goods and edges represent that two items are often purchased together. The node attributes are bag-of-words extracted from product reviews, and the product category determines the labels of nodes.

We use the pre-processed version provided by PyTorch-Geometric ~\cite{fey2019fast} for all datasets.

\subsection{Classifier}

For all the experiments, we follow the linear evaluation scheme as introduced in DGI ~\cite{velickovic2019deep}. Specifically, models is firstly trained with an unsupervised manner. After that, the learned representations are used to train and test a simple $l_2$-regularized logistic regression classifier with fixed learning rate and weight decay of $0.01$ and $0$ respectively.  

\subsection{Environment}
We implement our method with PyTorch and  PyTorch-Geometric~\cite{fey2019fast}, we adopt the Adam optimizer ~\cite{kingma2014adam} for parameter learning. All experiments are conducted on a NVIDIA GeForce RTX 3090 GPU with 24 GB memory.

\subsection{Hyper-parameter Settings} 
\label{sec:hyper}
\begin{table}[htbp]
  \centering
  \caption{Settings of major hyper-parameters}
    \begin{tabular}{ccccccc}
    \toprule[2pt]
    \multicolumn{2}{c}{Dataset} & epoch & $K$ Range & $K'$ Range & hidden\_size & proj\_size \\
    \midrule
    \multicolumn{2}{c}{Cora} & 500   & [0,4] & [1,4] & 512   & 512 \\
    \multicolumn{2}{c}{CiteSeer} & 400   & [2,4] & [1,3] & 512   & 512 \\
    \multicolumn{2}{c}{PubMed} & 900   & [0,3] & [0,3] & 512   & 128 \\
    \multicolumn{2}{c}{Coauthor-CS} & 1,000  & [0,3] & [1,3] & 256   & 256 \\
    \multicolumn{2}{c}{Amazon-P} & 1,500  & [3,5] & [1,3] & 256   & 256 \\
    \multicolumn{2}{c}{Amazon-C} & 2,000  & [1,4] & [0,3] & 256   & 256 \\

    \bottomrule[2pt]
    \end{tabular}%
  \label{tab:hyper1}%
\end{table}%

\begin{table}[htbp]
  \centering
  \caption{Settings of minor hyper-parameters}
    \begin{tabular}{ccccccccc}
    \toprule[2pt]
    \multicolumn{2}{c}{Dataset} & lr    & wd    & activation & $EDR_1$ & $EDR_2$ & $FDR_1$ & $FDR_2$ \\
    \midrule
    \multicolumn{2}{c}{Cora} & 2e-4 & 1e-06 & ReLU  & 0.3   & 0.3   & 0.3   & 0.3 \\
    \multicolumn{2}{c}{CiteSeer} & 1e-5 & 1e-06 & ReLU  & 0.3   & 0.2   & 0.3   & 0.2 \\
    \multicolumn{2}{c}{PubMed} & 0.002 & 1e-05 & ReLU  & 0.5   & 0.4   & 0.3   & 0.5 \\
    \multicolumn{2}{c}{Coauthor-CS} & 5e-4 & 1e-05 & RReLU & 0.3   & 0.2   & 0.3   & 0.4 \\
    \multicolumn{2}{c}{Amazon-P} & 0.001 & 0 & PReLU & 0.3   & 0.4   & 0.2   & 0.3 \\
    \multicolumn{2}{c}{Amazon-C} & 0.001 & 1e-05 & RReLU & 0.7   & 0.2   & 0.2   & 0.2 \\

    \bottomrule[2pt]
    \end{tabular}%
  \label{tab:hyper2}%
\end{table}%

We provide all the detailed hyper-parameters settings on the six datasets in Table \ref{tab:hyper1} and \ref{tab:hyper2}, where $EDR$ and $FDR$ are the edge dropping rate and feature dropping rate, respectively. Note that for the linear classifier in evaluation, we use the fixed learning rate of $0.01$ and weight decay of $0$ for fair comparisons. For most hyper-parameters, we follow the settings of GCA \cite{zhu2021graph} and all hyper-parameters are heuristically searched in the following space:
\begin{itemize}
\item The number of training epoches:$\{300, 400, 500, 800, 900, 1,000, 1,500, 2,000\}$
\item Range of $K$: $[low, high]$ for $low,high \in \{0,1,2,3,4,5\}$
\item Hidden size of $h$ operator: $\{64, 128, 256, 512\}$
\item Hidden size of projector: $\{64, 128, 256, 512\}$
\item Learning rate: $\{0.001, 0.002, 0.005, 1e-4, 2e-4, 5e-4, 1e-5, 2e-5, 5e-5\}$
\item Weight decay: $\{1e-4, 1e-5, 1e-6, 0\}$
\item Active function: $\{ReLU, PReLU, RReLU\}$
\item Edge dropping rate and feature dropping rate: $\{0.1, 0.2, 0.3, 0.4, 0.5, 0.6, 0.7\}$

\end{itemize}

\subsection{Graph diffusion}

Generalized graph diffusion was proposed by ~\cite{klicpera2019diffusion}, which generates a diffused and sparse version of the observed graph for different graph models and algorithms such as GNN and graph spectral clustering. MV-GCN~\cite{yuan2021semi} generates two different views with PPR and heat kernel and learns from both created views and the original view. Adaptive diffusion .\cite{zhao2021adaptive} is an extended version of graph diffusion that enables it to adaptively learn the neighborhood radius from data for observed graph and further try to generalize it for different feature channels and GNN layers. 

MVGRL ~\cite{hassani2020contrastive} combines graph contrastive learning and graph diffusion by treating graph diffusion as an efficient graph data augmentation method, which can naturally creates a different view of the original graph. Furthermore, MV-CGC~\cite{yuan2021semi} proposes a new graph contrastive learning framework with three views of original graph, diffusion graph and feature similarity view. However, those graph contrastive learning methods with graph diffusion focus to use graph diffusion as part of data augmentation, i.e. generating views, the focus of our approach is to perturb the encoder part to respectively help alleviate high-frequency noises, enrich training instances and bring safer augmentations.

\subsection{Node Clustering}

We adopt the k-means algorithm for node clustering, and use the implementation by kmeans-pytorch. We set the number of cluster centers of k-means algorithm as the number of classes in classification task and calculate the  normalized mutual information(NMI) to evaluate the results of clustering on six datasets. Clustering results are token the median over 20 randomly initialized runs and the results are report in ~\ref{tab:NMI} and MA-GCL can achieve SOTA performance on 5 out of 6 datasets.
\begin{table}[htbp]
  \centering
  \label{tab:NMI}%
  \caption{Performance of Node Clustering}
    \begin{tabular}{c|cccccc}
        \toprule[2pt]
    NMI   & Cora  & Citeseer & PubMed & Amazon-P & Amazon-C & Coauthor-CS \\
    \midrule
    raw feature & 0.144  & 0.199  & 0.178  & 0.261  & 0.194  & 0.563  \\
    Base Model & 0.471  & 0.429  & 0.221  & 0.535  & 0.474  & 0.601  \\
    GCA   & 0.474  & 0.441  & 0.224  & 0.587  & 0.499  & 0.614  \\
    S2GC  & 0.502  & 0.449  & 0.257  & 0.578  & 0.513  & 0.679  \\
    COLES & 0.470  & 0.415  & 0.262  & 0.546  & 0.497  & 0.660  \\
    ARIEL & 0.487  & 0.472  & 0.276  & 0.569  & 0.525  & 0.653  \\
    CCA-SSG & 0.517  & 0.476  & 0.254  & 0.591  & 0.517  & 0.677  \\
    MA-GCL & 0.509  & 0.494  & 0.297  & 0.602  & 0.533  & 0.696  \\
    \bottomrule[2pt]
    \end{tabular}%
    
\end{table}%

\subsection{Time and Memory Complexity}

We report both the time and memory usages with different $L$ of $L\in\{1, 2, 4, 6, 8, 10\}$ in ~\ref{tab:cost}, where $L=1$ denotes the base model. Note that the space complexity will not change as $L$ increases, and the time complexity is only sublinear to $L$. Taking the Coauthor-CS dataset as an example, the memory usages of $L=2$ and $L=10$ are both 7.5GB. And the training times of a single epoch are $0.053$s and $0.113$s, respectively. Thus the random strategy is very efficient.
\begin{table}[htbp]
  \centering
  \caption{Time and Memory Cost on Coauthor-CS Dataset}
    \begin{tabular}{ccccccc}
    \toprule[2pt]
    \#L   & 1     & 2     & 4     & 6     & 8     & 10 \\
    \midrule
    Memory(MiB) & 7446  & 7446  & 7446  & 7446  & 7446  & 7446 \\
    Time(s/epoch) & 0.045 & 0.053 & 0.069 & 0.083 & 0.097 & 0.113 \\
    \bottomrule[2pt]
    \end{tabular}%
  \label{tab:cost}%
\end{table}%

\subsection{Ablation Study}

We conduct ablation study to evaluate the effectiveness of our model augmentation strategies. Specifically, we compare the performance of four models of "Raw features", "Base model + MA", "Base model + DA" and "Base model + DA + MA". "Raw features" take the original node features as the inputs of classifier(logistic regression), "Base model + MA" only uses the model augmentation of our strategies to generate different view of embeddings, while "Base model + DA" adopts data augmentation of edges and features dropping, and "Base model + DA + MA" use both the model augmentation and data augmentation. According to the results in ~\ref{tab:ablation 2}, we can see that the performance of "Base model + MA" is better than "Raw features" and "Base model + MA + DA" is more effective than "Base model + MA", which demonstrates that our model augmentation strategies is useful and can be used with the data augmentation mechanism without conflict. 
\begin{table}[htbp]
  \centering
  \label{tab:ablation 2}%
  \caption{Ablation Study}
    \begin{tabular}{ccccccc}
    \toprule[2pt]
          & Cora  & Citeseer & PubMed & Amazon-P & Amazon-C & Coauthor-CS \\
\midrule
    raw feature & 47.9  & 49.3  & 69.1  & 78.5  & 73.8  & 90.4 \\
    Base Model +MA & 73.9  & 67.9  & 76.2  & 92.3  & 86.9  & 92.4 \\
    Base Model +DA & 81.1  & 71.4  & 79.1  & 91.1  & 87.7  & 92.9 \\
    Base Model +DA+MA & 83.3  & 73.6  & 83.5  & 93.8  & 88.8  & 94.2 \\
    \bottomrule[2pt]
    \end{tabular}%
\end{table}%

\subsection{Analysis of the Alignment Metric}

We conduct experiments to calculate the alignment metric mentioned in section 4.3 for MA-GCL and SimGRACE to prove that our model augmentation strategies bring safer augmentations. Specifically, we compute the mean value of the positive pair feature distances in MA-GCL and SimGRACE, and report the results in ~\ref{tab:metric align}.
\begin{table}[htbp]
  \centering
  \caption{Analysis of Alignment Metric}
    \begin{tabular}{ccccccc}
    \toprule[2pt]
          & Cora  & Citeseer & PubMed & Amazon-P & Amazon-C & Coauthor-CS \\
          \midrule
    Base Model +MA & 0.4577 & 0.3103 & 0.4531 & 0.2399 & 0.3017 & 0.2119 \\
    SimGRACE & 0.7199 & 0.5679 & 0.6921 & 0.7963 & 0.4762 & 0.3331 \\
    Base Model +DA & 0.3357 & 0.2459 & 0.5631 & 0.2013 & 0.3512 & 0.2258 \\
    Base Model +DA+MA & 0.5125 & 0.4063 & 0.6037 & 0.3347 & 0.3952 & 0.2701 \\
    \bottomrule[2pt]
    \end{tabular}%
  \label{tab:metric align}%
\end{table}%

\bibliographystyle{plain}
\bibliography{reference}